\documentclass[10pt,journal,compsoc]{IEEEtran}

\usepackage{amsmath}
\usepackage{diagbox}
\usepackage{graphics}
\usepackage{epsfig}
\usepackage{threeparttable}
\usepackage{xspace}
\usepackage{siunitx}
\usepackage{graphicx}
\usepackage{amssymb}
\usepackage{threeparttable}
\usepackage{wrapfig}
\usepackage{color}
\usepackage[normalem]{ulem}
\usepackage{multirow}
\usepackage{float}
\usepackage{amsfonts}
\usepackage{bm}
\usepackage{array}
\usepackage[table]{xcolor}
\usepackage{colortbl}
\usepackage{diagbox}
\usepackage{rotating}
\usepackage{booktabs}
\usepackage{overpic}
\usepackage{enumitem}
\usepackage{colortbl}
\usepackage{algorithm}
\usepackage{algorithmicx}
\usepackage{algpseudocode}
\usepackage{cite}
\usepackage[export]{adjustbox}
\usepackage{listings}
\usepackage{pifont}
\usepackage{subcaption}
\usepackage{tikz}
\usepackage{makecell}
\usepackage{arydshln}
\usepackage{tabu}
\usepackage[pagebackref=false,breaklinks=true,colorlinks,bookmarks=false]{hyperref}

\definecolor{mygray}{gray}{.92}

\definecolor{mygray}{gray}{.9}
\definecolor{ggray}{RGB}{127,127,127}
\definecolor{reda}{RGB}{192,0,0}
\definecolor{redb}{RGB}{217,148,143}
\definecolor{myyellow}{RGB}{190,144,0}
\definecolor{mygreen}{RGB}{80,100,40}
\definecolor{myblue}{RGB}{30,90,100}
\definecolor{mygray2}{gray}{.6}
\definecolor{mygray3}{gray}{.3}
\definecolor{mygray}{gray}{.9}
\definecolor{mywarning}{RGB}{233,144,61}
\definecolor{ggray}{RGB}{127,127,127}
\definecolor{reda}{RGB}{192,0,0}
\definecolor{redb}{RGB}{217,148,143}
\definecolor{myyellow}{RGB}{190,144,0}
\definecolor{mygreen}{RGB}{0,153,0}
\definecolor{mygreen2}{RGB}{153,255,153}
\definecolor{myred}{RGB}{177,35,15}
\definecolor{myblue}{RGB}{58,79,116}
\definecolor{myy}{RGB}{254,203,50}
\definecolor{myy2}{RGB}{191,144,0}
\definecolor{myg}{RGB}{205,205,205}
\definecolor{myg2}{RGB}{80,80,80}
\definecolor{codegreen}{RGB}{79,126,127}
\definecolor{codedefine}{RGB}{153,54,159}
\definecolor{codefunc}{RGB}{73,122,234}
\definecolor{codecall}{RGB}{73,122,234}
\definecolor{codepro}{RGB}{212,96,80}
\definecolor{codedim}{RGB}{89,152,195}

\definecolor{mygreen2}{RGB}{80,100,40}  

\usepackage{amsthm}
\theoremstyle{definition}

\usepackage{tabulary}
\newcolumntype{I}{!{\vrule width 1pt}}
\newcolumntype{x}[1]{>{\centering\arraybackslash}p{#1pt}}
\newcolumntype{y}[1]{>{\raggedright\arraybackslash}p{#1pt}}
\newcolumntype{z}[1]{>{\raggedleft\arraybackslash}p{#1pt}}

\definecolor{mygreen3}{HTML}{39b54a}  

\makeatletter
\newcommand{\thickhline}{%
	\noalign {\ifnum 0=`}\fi \hrule height 1pt
	\futurelet \reserved@a \@xhline
}
\makeatother

\makeatletter
\DeclareRobustCommand\onedot{\futurelet\@let@token\@onedot}
\def\@onedot{\ifx\@let@token.\else.\null\fi\xspace}

\def\eg{\emph{e.g}\onedot} 
\def\ie{\emph{i.e}\onedot} 
 
\def\etc{\emph{etc}\onedot} 
 
\def\etal{\emph{et al}\onedot}

\makeatother

\usepackage{cleveref}
\crefname{section}{§}{§§}

\newcommand{\added}[1]{\textcolor{red}{#1}}
\newcommand{\addedblue}[1]{\textcolor{blue}{#1}}
\newcommand{\addedbrown}[1]{\textcolor{brown}{#1}}

\renewcommand{\added}[1]{#1}
\renewcommand{\addedblue}[1]{#1}
\renewcommand{\addedbrown}[1]{#1}

\begin{document}
	\title{Self-supervised Adversarial Training of  Monocular Depth Estimation against Physical-World Attacks}
	\author{Zhiyuan Cheng, Cheng Han, James Liang, Qifan Wang, Xiangyu Zhang, Dongfang Liu
		\IEEEcompsocitemizethanks{
                \IEEEcompsocthanksitem Z. Cheng and X. Zhang are with Purdue University (Email: cheng443@purdue.edu, xyzhang@cs.purdue.edu)
                \IEEEcompsocthanksitem C. Han is with University of Missouri -- Kansas City and Rochester Institute of Technology (Email: chk9k@umsystem.edu; ch7858@rit.edu)
                \IEEEcompsocthanksitem D. Liu is with Rochester Institute of Technology (Email: dongfang.liu@rit.edu)
                \IEEEcompsocthanksitem J. Liang is with Rochester Institute of Technology and U.S. Naval Research Laboratory (Email: jcl3689@rit.edu)
                \IEEEcompsocthanksitem Q. Wang is with Meta AI (Email: wqfcr@fb.com)
			\IEEEcompsocthanksitem A preliminary version of this work has been published on ICLR 2023 as Spotlight~\cite{cheng2023adversarial}.
			\IEEEcompsocthanksitem Corresponding author: \textit{Dongfang Liu} and \textit{Xiangyu Zhang}
			%
		}
	}

	\markboth{IEEE TRANSACTIONS ON PATTERN ANALYSIS AND MACHINE INTELLIGENCE}%
	{Shell \MakeLowercase{\textit{et al.}}: Bare Demo of IEEEtran.cls for Journals}
	
	\IEEEtitleabstractindextext{
		\begin{abstract}
Monocular Depth Estimation (MDE) plays a vital role in applications such as autonomous driving. However, various attacks target MDE models, with physical attacks posing significant threats to system security. Traditional adversarial training methods, which require ground-truth labels, are not directly applicable to MDE models that lack ground-truth depth. Some self-supervised model hardening techniques (\eg, contrastive learning) overlook the domain knowledge of MDE, resulting in suboptimal performance. In this work, we introduce a novel self-supervised adversarial training approach for MDE models, leveraging view synthesis without the need for ground-truth depth. We enhance adversarial robustness against real-world attacks by incorporating $L_0$-norm-bounded perturbation during training. We evaluate our method against supervised learning-based and contrastive learning-based approaches specifically designed for MDE. Our experiments with two representative MDE networks demonstrate improved robustness against various adversarial attacks, with minimal impact on benign performance. Our code: \url{https://github.com/Bob-cheng/DepthModelHardening}.
		\end{abstract}
		
		}
	\maketitle
	
	\IEEEdisplaynontitleabstractindextext
	\IEEEpeerreviewmaketitle
	

\section{Introduction}\label{sec:intro}
 \IEEEPARstart{M}{onocular} Depth Estimation (MDE) is a deep neural network (DNN)-based task that estimates depth from a single image, allowing for 2D-to-3D projection by predicting the distance for each pixel in a 2D image\added{~\cite{ming2021deep}}. This makes it a cost-effective alternative to pricey Lidar sensors. Applications of MDE are vast, including autonomous driving~\cite{tesla-self-supervised}, visual SLAM~\cite{wimbauer2020monorec}, and visual relocalization~\cite{lm-reloc-2020}, \etc. Specifically, self-supervised MDE has become increasingly popular in the industry (\eg, Tesla Autopilot~\cite{tesla-self-supervised}) due to its ability to achieve comparable accuracy without requiring ground-truth depth data from Lidar during training. However, \added{due to the known vulnerabilities in DNNs~\cite{goodfellow2014explaining,cheng2024fusion}}, several digital-world~\cite{zhang2020adversarial,wong2020targeted,cheng2024badpart} and physical-world adversarial attacks~\cite{cheng2022physical} has been developed against MDE. These attacks primarily use optimization-based methods to create adversarial input that deceive the MDE models. Given the significance and widespread use of MDE models, these adversarial attacks pose a substantial risk to the security of applications such as autonomous driving. As a result, there is an urgent need to develop and strengthen MDE models against these threats. 

Adversarial training~\cite{goodfellow2014explaining} is a widely recognized and effective defense against adversarial attacks, \added{which hardens target models by exposing them to adversarial examples during the training process.} However, it typically necessitates ground-truth labels during training, making it unsuitable for hardening MDE models that lack depth ground truth. Although contrastive learning has recently garnered significant attention and has been employed for self-supervised adversarial training~\cite{ho2020contrastive,kim2020adversarial}, it does not take into account the domain knowledge of depth estimation, resulting in suboptimal outcomes (as demonstrated in \S\ref{sec:main_results}). Furthermore, many existing adversarial training methods do not account for certain characteristics of real-world attacks\added{~\cite{cheng2022physical,eykholt2018robust}, such as adversarial patches with intense perturbations and perspective variations of target objects}. Therefore, in this paper, we concentrate on the challenge of \textit{strengthening MDE models against physical-world attacks using self-supervised adversarial training without the need for ground-truth depth}.

\begin{figure*}[th]
    \centering
    \includegraphics[width=0.9\textwidth]{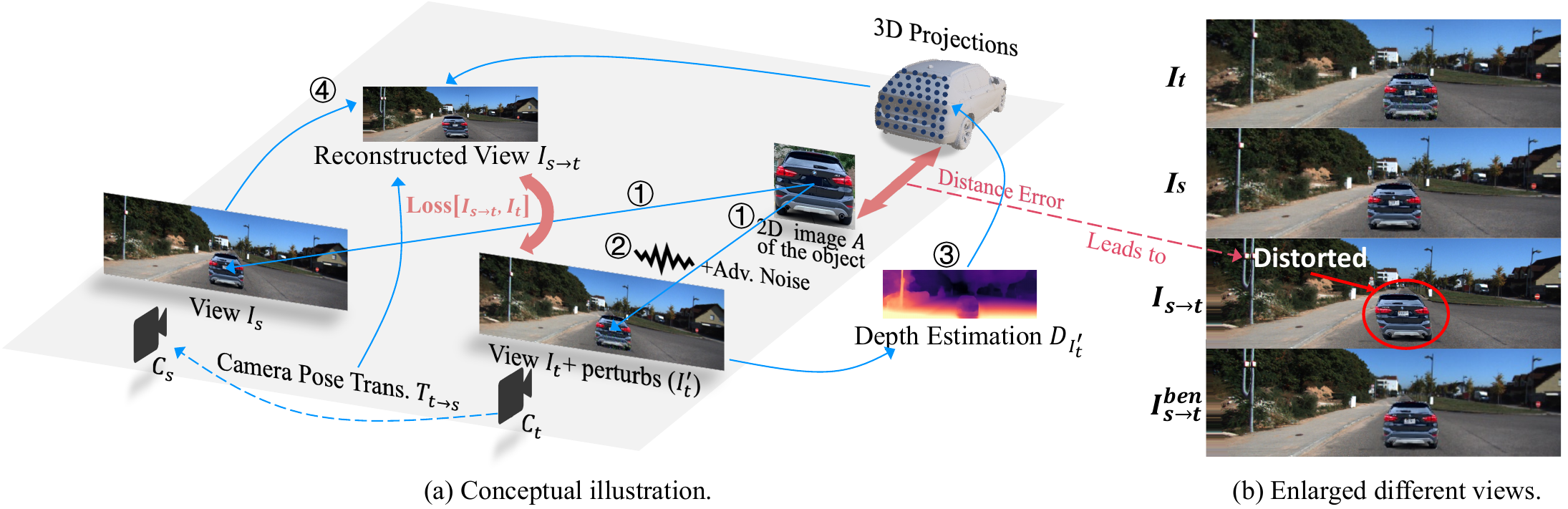}
    \caption{\small \added{Self-supervised adversarial training of MDE with view synthesis.} }
     \label{fig:highlight_intro}
     \vspace{-12pt}
\end{figure*}

A straightforward proposal for enhancing MDE models is to perturb 3D objects across various scenes while ensuring that the estimated depths remain accurate. However, implementing such adversarial training is challenging. Firstly, 3D perturbations are difficult to achieve in the physical world. Even if a model is trained in simulation, a high-fidelity simulator and a powerful scene rendering engine are required to accurately project 3D perturbations into 2D variations. Secondly, since self-supervised MDE training lacks ground-truth depth, even if realistic 3D perturbations could be obtained and utilized in training, the model might converge on incorrect (but robust) depth estimations.

In this paper, we introduce a novel self-supervised adversarial training approach for MDE models. Figure~\ref{fig:highlight_intro}a offers a conceptual depiction of our method. A board $A$ displaying the 2D image of a 3D object (\eg, a car) is positioned at a fixed location (next to the car at the top-right corner). Two cameras (close to each other at the bottom) $C_t$ and $C_s$ provide \added{two adjacent views (\ie stereo view~\cite{Stereoscopy})} of the board (images $I_t$ and $I_s$ in Figure~\ref{fig:highlight_intro}b). There are fixed geometric relationships between pixels in the two 2D views captured by each camera, allowing the image from one view to be transformed into the image from the other view. \added{Intuitively, the target view image $I_t$ can be obtained by shifting the source view image $I_s$ to the right. If two cameras are unavailable, two temporally adjacent frames in a video stream can be used to form the two views. }

\addedbrown{During training, four main steps are taken. \textbf{\large\ding{192}}: Camera $C_t$ captures an image $I_t$ of the original 2D image board $A$, and camera $C_s$ takes a picture $I_s$ of the board as well. 
\textbf{\large\ding{193}}: Adversarial noise is applied to $A$, aiming to deceive the target MDE model into overestimating its depth, resulting in a perturbed view $I'_t$. 
\textbf{\large\ding{194}}: The perturbed image $I_t$+perturs ($I'_t$) is input into the MDE model for depth estimation, then we project the estimated distance $D_{I'_t}$ to 3D space. The blue dots on the car denote the projected pixels of the target object. 
Due to the perturbations, a susceptible model generates distance errors, as indicated by the red arrow between $A$ and the projected pixels in Figure~\ref{fig:highlight_intro}a. \textbf{\large\ding{195}}:~Given the relative pose transformations $T_{t\rightarrow s}$ between the two cameras, we map the projected pixels in 3D space back onto camera $C_s$'s view to locate the corresponding pixels in the adjacent image $I_s$, then we attempt to reconstruct target image $I_t$ from source image $I_s$ by rearranging the pixels.}

Owing to the distance error, the reconstructed image $I_{s\rightarrow t}$ (shown in Figure~\ref{fig:highlight_intro}b) differs from $I_t$, causing part of the car (the upper part inside the red circle) to be distorted. In contrast, Figure~\ref{fig:highlight_intro}b also displays the reconstructed image $I^{ben}_{s\rightarrow t}$ without the perturbation, which closely resembles $I_t$. The objective of our training (for the target MDE model) is to minimize the differences between original and reconstructed images. The above process is conceptual and would require significant physical-world overhead to be realized faithfully. In \S\ref{sec:methods}, we describe how to circumvent the majority of the physical-world costs through image synthesis and training on synthesized data.

Traditional adversarial training often assumes bounded perturbations in $L_2$ or $L_{\infty}$ norm (i.e., measuring the overall perturbation magnitude across all pixels), while real-world attacks are typically unbounded in these norms. These attacks tend to be stronger in order to persist amid varying environmental conditions. To strengthen MDE models against such attacks, we employ a loss function that can efficiently approximate the $L_0$ norm (measuring the number of perturbed pixels regardless of their perturbation magnitude) while remaining differentiable. Adversarial samples produced by minimizing this loss can effectively simulate physical attacks. \added{Also, the ``board" ($A$ in Figure~\ref{fig:highlight_intro}a) serves as a pragmatic approximation of a 3D model of the target object. This simplification, using a 2D board, is both realistic and practical, since adversarial patches are commonly affixed to flat surfaces in real-world attacks. We manipulate the board with random distances and viewing angles during training to simulate perspective shifts encountered in real-world patch attacks, resulting in a more robust defense. Moreover, this approach significantly curtails the training overhead in view synthesis and boosts efficiency.}

\addedbrown{We evaluate our approach and compare it with a supervised learning baseline and a contrastive learning baseline, adapted from state-of-the-art adversarial contrastive learning~\cite{kim2020adversarial}. Our results demonstrate that our method achieves superior robustness against various adversarial attacks, with minimal degradation in model performance. The average depth estimation error for an adversarial object with 1/10 area of perturbation is reduced from 6.08 m to 0.25 m using our technique, outperforming the 1.18 m error from the supervised learning baseline. Furthermore, the contrastive learning baseline significantly degrades model performance. A video of our real-world experiments is available at \url{https://youtu.be/_b7E4yUFB-g}.} Our main contributions are as follows:


\begin{itemize}[leftmargin=10pt]
  \item We develop a novel technique for synthesizing 2D images that adhere to real-world constraints (\eg, relative camera positions), and directly apply perturbations to these images during adversarial training. This approach significantly reduces the physical world costs involved.
  
  \item Our approach leverages {\em the reconstruction consistency} between two different views to facilitate self-supervised adversarial training without ground-truth depth labels.
  
  \item We generate $L_0$-bounded perturbations using a differentiable loss function and randomize camera and object settings during the synthesis process to simulate real-world attacks and enhance the robustness of the model.
\end{itemize}

This work significantly extends our conference paper~\cite{cheng2023adversarial} in various aspects:

\addedbrown{(1)} We broaden our approach to include a general training framework that can be applied to any monocular depth estimation method against physical-world attacks (\S\ref{sec:methods}) and provide more adversarial attack examples and qualitative defense results to evaluate the performance (\S\ref{append:adv_egs}).
    
\addedbrown{(2)} To ensure the effectiveness of view synthesis, we enhance the view synthesis process by considering inconsistent environmental lighting between the scene and object and perform additional ablative experiments on the impact of varying viewing angles during training, demonstrating the strength of our view synthesis algorithm (\S\ref{app:ang_range}). 

\addedbrown{(3)} We conduct a human evaluation to assess the quality and reality of our synthesized images (\S\ref{app:syn-quality}), and investigate the effects of inaccurate projections and \added{unrealistic object positions (\S\ref{app:inacc_proj})}.

\addedbrown{(4)} We present additional transferability evaluation (\S\ref{append:target_objs}) for our hardened models applied to other target objects and validate models' robustness against a wider range of adversarial attacks including black-box ones (\S\ref{app:more_atks}).
    
\addedbrown{(5)} We compare the differences between fine-tuning and training from scratch (\S\ref{app:scratch}), explore the model performance when combining various methods (\S\ref{app:combi}), and contrast the supervised baseline trained with pseudo-depth labels and ground-truth depth labels (\S\ref{app:sup_GT}).
    
\addedbrown{(6)} We extend our method to indoor scenes and more advanced MDE networks such as Manydepth (\S\ref{app:indoor-manydepth}). 

\addedbrown{(7)} We present an in-depth analysis of our method's limitations and potential in \S\ref{app:limit}, which we hope will inspire further research collaboration.	
\vspace{-10pt}
\section{Related Work}%
This section provides an overview of the most pertinent research on Self-supervised MDE (\S\ref{Self-supervised MDE}), MDE Attack and Defense (\S\ref{MDE Attack and Defense}), and Adversarial Robustness (\S\ref{Adversarial Robustness}).

\vspace{-10pt}
\subsection{Monocular Depth Estimation} \label{Self-supervised MDE}
Due to the benefits of training without depth ground truth, MDE has recently attracted significant interest in a self-supervised manner. In this kind of training approach, monocular videos and/or stereo image pairs serve as input. Essentially, two images captured by a camera or cameras from neighboring positions are used in each optimization iteration. A depth network and a pose network are employed to estimate the depth map of one image and the transformation between the two camera poses, respectively. Using the depth map and pose transformation, pixel correspondence across the images is calculated, and then an attempt is made to rearrange the pixels in one image to reconstruct the other. Both the pose network and depth network are updated simultaneously to minimize reconstruction error. \added{Please refer to \cite{zhou2017unsupervised} for a detail explanation of this process.} Garg \etal~\cite{garg2016unsupervised} first proposed using color consistency loss between stereo image pairs in training. Zhou \etal~\cite{zhou2017unsupervised} enabled video-based training with two networks (one depth network and one pose network). Many subsequent works improved self-supervision with new loss terms~\cite{godard2017unsupervised,bian2019unsupervised,wang2018learning,yin2018geonet,ramamonjisoa-2021-wavelet-monodepth,yang2020d3vo,lu2023transflow} or incorporated temporal information~\cite{wang2019recurrent,zou2020learning,tiwari2020pseudo,watson2021temporal}. Among them, Monodepth2~\cite{monodepth2} significantly enhanced performance with several innovative designs, such as minimum photometric loss selection, masking out static pixels, and multi-scale depth estimation. Depthhints~\cite{watson2019self} further improved upon Monodepth2 by incorporating additional depth suggestions obtained from stereo algorithms. While unsupervised training has proven effective, enhancing its robustness against physical attacks remains an open challenge.

\begin{figure*}[t]
    \centering
        \includegraphics[width=0.6\textwidth]{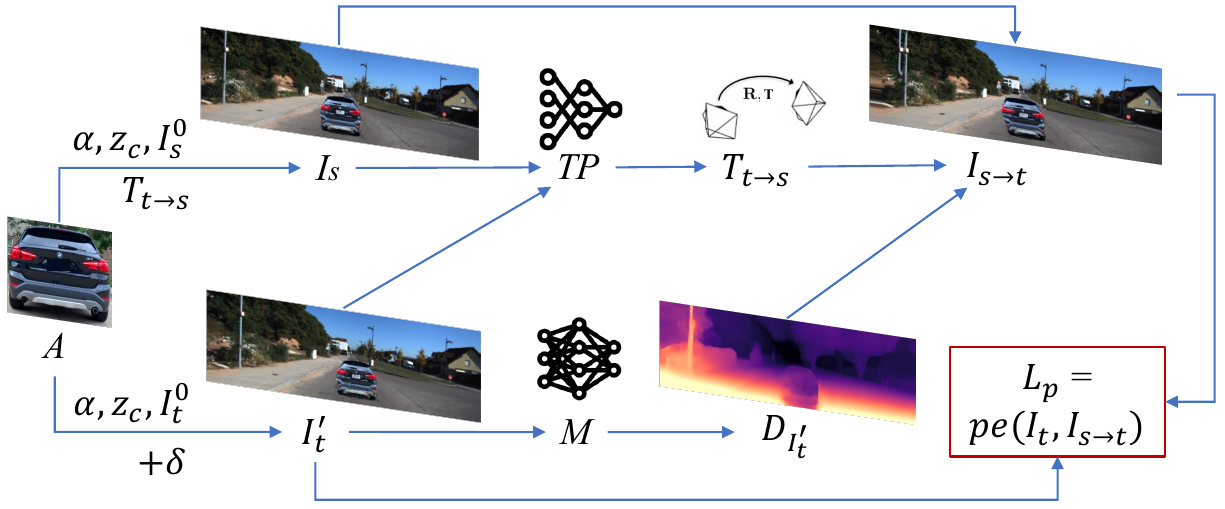}
        \caption{\small \added{The pipeline of self-supervised adversarial training of monocular depth estimation.}
        }
        \label{fig:adv_training}
        \vspace{-12pt}
\end{figure*}

\vspace{-10pt}
\subsection{MDE Attack and Defense} \label{MDE Attack and Defense}
Mathew \etal~\cite{mathew2020monocular} employed a deep feature annihilation loss to launch perturbation and patch attacks. Zhang \etal~\cite{zhang2020adversarial} designed a universal attack using a multi-task strategy, while Wong \etal~\cite{wong2020targeted} generated targeted adversarial perturbations on images, which can arbitrarily alter the depth map. Hu \etal~\cite{hu2019analysis} proposed a defense method against perturbation attacks by masking out non-salient pixels, requiring an additional saliency prediction network. Regarding physical-world attacks, Cheng \etal~\cite{cheng2022physical} created a printable adversarial patch to make a vehicle disappear. To the best of our knowledge, our work is the first to focus on enhancing the robustness of MDE models against physical-world attacks.

\vspace{-10pt}
\subsection{Adversarial Robustness} \label{Adversarial Robustness}
Deep neural networks (DNNs) are known to be vulnerable to adversarial attacks, \added{which are usually optimized perturbations on the model input that could lead to model misbehavior~\cite{szegedy2013intriguing,madry2018towards,moosavi2016deepfool}. 
Adversarial training is a popular defensive approach that can harden DNN models against adversarial attacks~\cite{madry2018towards,carlini2017towards,tramer2017ensemble}. The core concept involves integrating adversarial examples into the training process by augmenting the training data, thereby pre-exposing the model to adversarial noise and improving the run-time robustness. Typically, during adversarial training, the process alternates between conducting model training to minimize the training loss and generating adversarial examples that can maximize the loss.} This approach has been applied to various domains, including image classification~\cite{carlini2017towards,madry2018towards}, object detection~\cite{zhang2019towards,chen2021robust,chen2021class}, and segmentation~\cite{xu2021dynamic,hung2018adversarial,arnab2018robustness}, \etc.
However, adversarial training often requires supervision, as generating adversarial examples necessitates ground truth, and most tasks demand labels for training. Some semi-supervised adversarial learning methods~\cite{carmon2019unlabeled,alayrac2019labels} use a small portion of labeled data to enhance robustness. Contrastive learning~\cite{ho2020contrastive,kim2020adversarial} is also employed with adversarial examples for self-supervised learning. In this work, we investigate self-supervised adversarial training for MDE without ground-truth depth and compare our method with contrastive learning-based and supervised learning-based methods for MDE.
In the realm of autonomous driving, there are other works focusing on attacking Lidar or multi-sensor fusion-based systems\cite{tu2020physically,tu2021exploring,cao2019adversarial,cao2021invisible}. These approaches use sensor spoofing or adversarial shapes to deceive Lidar hardware or AI models. In contrast, our work considers fully vision-based systems in which MDE is the key component.

\vspace{-10pt}
\section{Methodology}\label{sec:methods}

Our approach is composed of several components. \addedbrown{The first component (\S\ref{sec:MDE_training}) deals with self-supervised adversarial training pipeline and the view reconstruction process (steps \ding{195} in Figure~\ref{fig:highlight_intro}a).} The \addedbrown{second} component (\S\ref{sec:view_syn}) involves view synthesis, which generates two views $I_t$ and $I_s$ of the object, corresponding to step \ding{192} in Figure~\ref{fig:highlight_intro}a. The \addedbrown{third} component (\S\ref{sec:adv_perturb}) focuses on robust adversarial perturbation, altering $I_t$ to produce maximum distance errors (step \ding{193} in Figure~\ref{fig:highlight_intro}a). \addedbrown{We discuss the specifics of these components as follows.}

\vspace{-10pt}
\subsection{Self-supervised MDE Training}\label{sec:MDE_training}

\added{Figure~\ref{fig:adv_training} depicts an overview of our self-supervised adversarial training pipeline. From left to right, the pipeline starts with the object image $A$. First, we synthesize it onto two adjacent street view backgrounds, $I^0_s$ and $I^0_t$, sampled from the dataset to create the two views of the object, $I_s$ and $I_t$. During synthesis, we use various distances $z_c$ and viewing angles $\alpha$. The synthesis process circumvents physical-world costs significantly, and we explain the details in \S\ref{sec:view_syn}. Next, robust adversarial perturbations $\delta$ are generated onto $A$ while synthesizing $I_t$, forming the adversarial sample $I'_t$, which is then fed into the depth network $M$ to obtain depth $D_{I'_t}$. This process is detailed in \S\ref{sec:adv_perturb}. Recall that we need the camera pose transformation matrix $T_{t\rightarrow s}$ between the coordinate systems of the two cameras. We predict it by a camera transposing model {\it TP} with the two street views $I^0_s$ and $I^0_t$ as input.} \addedbrown{This TP network is adapted from Monodepth2~\cite{monodepth2}. Its architecture is formed from a ResNet18, modified to accept a pair of color images (or six channels) as input, and to predict a single 6-DoF (Degrees of Freedom) relative pose. For a comprehensive understanding of the model, we refer readers to the original publication.} \added{Next, we reconstruct a version of the target view $I_t$ from the source view $I_s$, using $D_{I'_t}$ and the output $T_{t\rightarrow s}$ of the {\it TP} network. We refer to the resultant image as $I_{s\rightarrow t}$. Intuitively, $I'_t$ induces depth errors which distort the reconstruction from $I_s$ to $I_{s\rightarrow t}$, making the latter appear different from $I_t$. At last, we use a loss $L_p$ of the reconstruction error to train the two models $M$ and $TP$.}

Specifically, the reconstruction of the target $I_t$ is achieved by utilizing the pixel-to-pixel correspondence between $I_s$ and $I_t$. The function for projecting a pixel $(u^t, v^t)$ in image $I_t$ to a pixel $(u^s, v^s)$ image $I_s$ is defined as follows:
\begin{gather}\label{eq:reconstruction}
    \begin{aligned}
        \begin{bmatrix}
            x^{t}~y^{t}~z^{t}~1
        \end{bmatrix}^\top &= D_{I'_t}[u^{t},v^{t}] \cdot  K^{-1}\cdot \begin{bmatrix}
            u^{t}~v^{t}~1
        \end{bmatrix}^\top,\\ 
         \begin{bmatrix}
            x^s~y^s~z^s~1
        \end{bmatrix}^\top &= T_{t\rightarrow s} \cdot \begin{bmatrix}
            x^{t}~y^{t}~z^{t}~1
        \end{bmatrix}^\top,\\
        \begin{bmatrix}
            u^s~v^s~1
        \end{bmatrix}^\top &= 1/z^s\cdot K\cdot \begin{bmatrix}
            x^s~y^s~z^s~1
        \end{bmatrix}^\top.
    \end{aligned}
\end{gather}
Conceptually, there exist relationships between 2D image pixels and 3D coordinates, as illustrated by the first and third formulas in Equation~\ref{eq:reconstruction}. The 3D coordinates also have correlations determined by camera poses, as demonstrated by the second formula. \added{Basics of the camera projections can be found at \cite{camera-resectioning}.} It is important to note that the first 2D-to-3D relationship, corresponding to step {\large \ding{194}} in Figure~\ref{fig:highlight_intro}a, is parameterized on $D_{I'_t}$, which is the depth estimation of $I'_t$ .
Let $\begin{bmatrix}
            u^s~v^s
        \end{bmatrix} = P_{D_{I'_t},T_{t\rightarrow s}}(u^{t}, v^{t})$
        be the transformation function that maps a pixel in $I_{t}$ to a pixel in $I_s$, derived from Equation~\ref{eq:reconstruction}.
$I_{s\rightarrow t}$ is reconstructed as:
\begin{gather}
    \begin{aligned}\label{eq:target_recons}
        I_{s\rightarrow t}[u, v] = I_s[ P_{D_{I'_t},T_{t\rightarrow s}}(u, v)].
    \end{aligned}
\end{gather}
Intuitively, this method reorganizes the pixels in $I_s$ to create $I_{s\rightarrow t}$. 
$I_{s\rightarrow t}$ is then compared with $I_t$, and the core training objective for our approach is defined as follows:
\begin{align}\label{eq:recons_loss}
    \min\limits_{\theta_{M}, \theta_{\textit{TP}}} \mathcal{L}_p &= pe(I_t, I_{s\rightarrow t}),
\end{align}
which is to adjust the weight values of $M$ and {\it TP} to minimize the photometric reconstruction error, represented by $pe()$. \addedbrown{The specific design of $pe()$ varies in the literature, and we use the same one as Monodepth2~\cite{monodepth2} in our experiments. Nevertheless, alternative formulations of $pe()$ could conceivably be integrated within the framework of our model hardening approach.}


\begin{figure*}[t]
    \centering
    \includegraphics[width=0.8\textwidth]{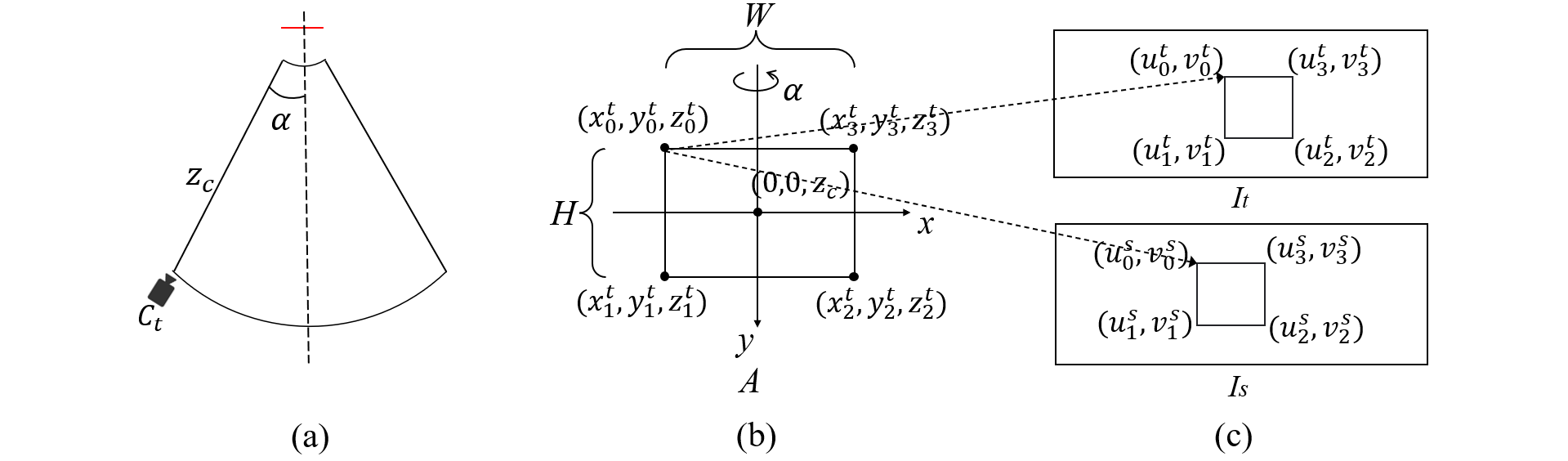}
    \caption{\small (a) A top-down bird view of the relative positions of the camera and the target object. (b) The 3D coordinates of the object's four corners in the camera frame. (c) Projection of the physical-world object onto the two views. }\label{fig:synth_def}
    \vspace{-12pt}
\end{figure*}

\vspace{-8pt}
\subsection{View Synthesis To Avoid Physical Scene Mutation}\label{sec:view_syn}
As discussed in \S\ref{sec:intro}, conceptually, we need to place an image board of the 3D object at various physical locations on streets and use two cameras to capture images of the board. To achieve robustness during training, we must vary the object's image (\eg, different vehicles), the position of the image board, the camera positions and viewing angles, and the street view. This would involve significant overhead in the physical world. Consequently, we propose an innovative view synthesis technique that minimizes physical world overhead. Instead, it directly synthesizes $I_s$ and $I_t$ by incorporating an object into some street views, considering different settings of the aforementioned configurations.

In particular, we consider a camera $C_t$ and a board attached with an object image $A$ in the physical world. The board's physical width and height are $W$ and $H$, respectively. As illustrated in Figure~\ref{fig:synth_def} (b), the four corners of the board have the 3D coordinates ($x^t_0$, $y^t_0$, $z^t_0$), ..., and ($x^t_3$, $y^t_3$, $z^t_3$) in camera $C_t$'s coordinate system with the camera as the origin. The board is placed at a distance $z_c$ from the camera, with an angle of $\alpha$ (see Figure~\ref{fig:synth_def} (a)). The board's size is true to the rear of the actual object, which is crucial for realistic physical-world view synthesis. Subsequently, we establish a projection function, defined by variables such as $W$, $H$, $z_c$, $\alpha$, \etc, that enables the transformation of a pixel from $A$ to a pixel in the view $I_t$ of camera $C_t$. This function includes a 3D-to-2D projection process, which is designed in such a way that by adjusting $z_c$ and $\alpha$, we can generate numerous instances of $I_t$ with varying perspectives of $A$.

Obtaining $I_s$, which is intended to construct a stereo pair with $I_t$, does not require a second physical camera. Instead, we utilize two successive frames from a street view video, such as those found in the KITTI dataset~\cite{Geiger2013IJRR}, denoted as $I^0_s$ and $I^0_t$. These frames can mimic a stereo pair captured by two adjacent cameras. The distinctions between the two images inherently denote the relative positions of the two cameras. A significant advantage of this approximation is the ease with which we can simulate a wide range of camera placements and street views by selecting different consecutive video frames. This is in line with existing research~\cite{monodepth2,watson2019self}. We substitute the portion of $I_t$ that does not belong to $A$, i.e., the object's background, with the corresponding part in $I^0_t$. In essence, we create a realistic $I_t$ by overlaying the image of the object onto a background image $I^0_t$, while maintaining respect for physical world constraints. 

\addedbrown{The adjacent view $I_s$ with $A$ is also synthesized using $I^0_s$ in a similar way, but this time, we need to consider the camera pose transformation from $C_t$ to $C_s$. This transformation, represented by $T_{s\rightarrow t}$ and established from both $I^0_s$ and $I^0_t$, converts the 3D coordinates of $A$ in the coordinate system of camera $C_t$ to that of camera $C_s$. Then, the same 3D-to-2D projection is utilized to synthesize $I_s$ (Figure~\ref{fig:synth_def}(c)). Consequently, the resulting view of $A$ in $I_s$ aligns with the camera pose represented by $I^0_s$.
$I_t$ and $I_s$ are then incorporated into model hardening. }

Formally, when the center of camera $C_t$'s view is in alignment with the center of the image board $A$ (see Figure~\ref{fig:synth_def}), the relationship between a pixel ($u^A,v^A$) in $A$ and its corresponding 3D coordinate ($x^t$, $y^t$, $z^t$) can be defined as:
\begin{gather}\label{eq:img2board}
\scalebox{0.8}{$
    \begin{aligned}
        \begin{bmatrix}
            x^t\\y^t\\z^t\\1
        \end{bmatrix} = 
        \begin{bmatrix}
            \cos\alpha & 0 & -\sin\alpha & 0\\
            0 & 1 & 0 & 0\\
            \sin\alpha & 0 & \cos\alpha & z_c \\
            0 & 0 & 0 & 1
        \end{bmatrix}\cdot
        \begin{bmatrix}
            W/w & 0 & -W/2\\
            0 & H/h & -H/2\\
            0 & 0 & 0\\
            0 & 0 & 1
        \end{bmatrix}\cdot
        \begin{bmatrix}
            u^A \\ v^A \\ 1
        \end{bmatrix},
    \end{aligned}
$}
\end{gather}
where $w$ and $h$ represent the pixel dimensions (\ie, width and height, respectively) of $A$. The remaining variables (\eg, $\alpha$, $z_c$) are explained in Figure~\ref{fig:synth_def}. The 3D coordinates can then be further transformed into pixels in $I_t$ and $I_s$ as follows:
\begin{gather}\label{eq:pixel_proj}
    \begin{aligned}
        \begin{bmatrix}
            u^t~v^t~1
        \end{bmatrix}^\top &= 1/z^t\cdot K\cdot 
        \begin{bmatrix}
            x^t~y^t~z^t~1
        \end{bmatrix}^\top,\\
        \begin{bmatrix}
            x^s~y^s~z^s~1
        \end{bmatrix}^\top &= T_{t\rightarrow s} \cdot\begin{bmatrix}
            x^t~y^t~z^t~1
        \end{bmatrix}^\top,\\
        \begin{bmatrix}
            u^s~v^s~1
        \end{bmatrix}^\top &= 1/{z^s}\cdot K \cdot
        \begin{bmatrix}
            x^s~y^s~z^s~1
        \end{bmatrix}^\top,
    \end{aligned}
\end{gather}
$K$ represents the camera's intrinsic parameters\added{~\cite{camera-resectioning}}.
By consolidating Equation~\ref{eq:img2board} with Equation~\ref{eq:pixel_proj}, we can formulate the projections from a pixel $(u^A, v^A)$ of the object image to a pixel $(u^t, v^t)$ in $I_t$ and to a pixel $(u^s, v^s)$ in $I_s$.
Let     $\begin{bmatrix}
            u^t~v^t~1
        \end{bmatrix}^\top = P^{A\rightarrow t}_{z_c, \alpha}(u^A, v^A)$ and $\begin{bmatrix}
            u^s~v^s~1
        \end{bmatrix}^\top = P^{A\rightarrow s}_{z_c, \alpha, T_{t\rightarrow s} }(u^A, v^A)$,
we synthesize $I_t$ and $I_s$:
\begin{gather}
\begin{aligned}\label{eq:it_syn}
    I_t[u,v]=\left\{ 
        \begin{aligned}
          A[u^A,v^A],  &  \begin{bmatrix}
            u~v~1
        \end{bmatrix}^\top = P^{A\rightarrow t}_{z_c, \alpha}(u^A, v^A) \\
        I^0_t[u,v],  & \quad \qquad \qquad \text{otherwise}
        \end{aligned}\right.,
\end{aligned}
\end{gather}
\begin{gather}
\begin{aligned}\label{eq:is_syn}
     I_s[u,v]=\left\{
        \begin{aligned}
          A[u^A,v^A],  &  \begin{bmatrix}
            u~v~1
        \end{bmatrix}^\top = P^{A\rightarrow s}_{z_c, \alpha, T_{t\rightarrow s} }(u^A, v^A) \\
        I^0_s[u,v],  & \quad \quad \qquad \qquad \text{otherwise}
        \end{aligned}\right.,
\end{aligned}
\end{gather}
where $I^0_t$ and $I^0_s$ are the background images that indirectly represent the relative positions of the cameras. By adjusting variables such as $I^0_t$, $I^0_s$, $z_c$, $\alpha$, $A$, we can generate a vast number of $I_t$ and $I_s$ instances which are then utilized in the hardening process. The production of this synthetic data incurs practically no expense compared to generating a comparably diverse dataset from the physical world.


\begin{table*}[t]
    \centering
    \caption{\small \textbf{Benign performance} of original and hardened models on depth estimation. }
    \label{tab:ben_perf}
    \vspace{-5pt}
    \scalebox{1.1}{
    \begin{threeparttable}
    \begin{tabular}{lcccccccccc}
\toprule
  & \multicolumn{5}{c}{Monodepth2} & \multicolumn{5}{c}{DepthHints} \\ \cline{2-11} Models
 & ABSE$\downarrow$ & RMSE$\downarrow$ & ABSR$\downarrow$ & SQR$\downarrow$ & $\delta\uparrow$ & ABSE$\downarrow$ & RMSE$\downarrow$ & ABSR$\downarrow$ & SQR$\downarrow$ & $\delta\uparrow$ \\ \midrule\midrule
Original & 2.125 & 4.631 & 0.106 & 0.807 & 0.877 & 2.021 & 4.471 & 0.100 & 0.728 & 0.886 \\
L0+SelfSup (Ours) & 2.16 & 4.819 & 0.105 & 0.831 & 0.874 & 2.123 & 4.689 & 0.103 & 0.777 & 0.877 \\
L0+Sup & 2.162 & 4.648 & 0.110 & 0.846 & 0.876 & 2.015 & 4.453 & 0.100 & 0.734 & 0.887 \\
L0+Contras & 3.218 & 6.372 & 0.155 & 1.467 & 0.782 & 3.626 & 6.742 & 0.209 & 1.561 & 0.694 \\
PGD+SelfSup & 2.169 & 4.818 & 0.105 & 0.826 & 0.874 & 2.120 & 4.680 & 0.103 & 0.774 & 0.877 \\
PGD+Sup & 2.153 & 4.637 & 0.109 & 0.838 & 0.876 & 2.019 & 4.460 & 0.101 & 0.736 & 0.886 \\
PGD+Contras & 3.217 & 6.083 & 0.194 & 1.825 & 0.756 & 3.928 & 7.526 & 0.213 & 2.256 & 0.701 \\ \bottomrule
\end{tabular}
\begin{tablenotes}
    \item  * For$_{\!}$ hardened$_{\!}$ models,$_{\!}$ A+B$_{\!}$ denotes$_{\!}$ generating$_{\!}$ adversarial$_{\!}$ perturbation$_{\!}$ with$_{\!}$ method$_{\!}$ A$_{\!}$ and$_{\!}$ training$_{\!}$ with$_{\!}$ method$_{\!}$ B.
\end{tablenotes}
\end{threeparttable}
}
\vspace{-10pt}
\end{table*}

\vspace{-10pt}
\subsection{Robust Adversarial Perturbations}\label{sec:adv_perturb}
We employ an optimization-based method to create robust adversarial perturbations, denoted as $\delta$, on the object image $A$. This results in the corresponding adversarial object $A+\delta$, and we synthesize $I'_t$ by substituting $A$ with $A+\delta$ in Equation~\ref{eq:it_syn}. The synthesized $I'_t$ is subsequently utilized in adversarial training.
We limit the perturbations using the $L_0$-norm, which effectively controls the number of perturbed pixels. This is a departure from digital-world attacks that typically use $L_\infty$-norm or $L_2$-norm-bounded perturbations, like FGSM~\cite{goodfellow2014explaining}, Deepfool~\cite{moosavi2016deepfool}, and PGD~\cite{madry2018towards}. 

Physical-world attacks usually employ adversarial patches~\cite{brown2017adversarial} with no restrictions on the magnitude of perturbations within the patch area. This is because larger perturbations are often required to cause consistent model errors in the face of diverse environmental conditions, such as changes in lighting, viewing angles, distance, and camera noise.
Therefore, $L_0$-norm is a better fit for physical-world attacks as it restricts the number of pixels to be perturbed without limiting the perturbation magnitude of individual pixels. However, the calculation of $L_0$-norm is non-differentiable by definition, making it unsuitable for optimization. For this reason, we adopt a `soft' version of it, as proposed in~\cite{Tao2022BetterTrigger}. The core idea is to separate the perturbations into positive or negative components and utilize the long-tail effects of the $\tanh$ function in the normalization term. This effectively models the two extremes of a pixel's value change: zero perturbation or arbitrarily large perturbation. As a result, a pixel tends to either have a large perturbation or no perturbation at all. 
\begin{equation}
        \delta = ~maxp \cdot (\texttt{clip}(b_p, 0, 1) - \texttt{clip}(b_n, 0, 1))  \label{eq:perturb}.
\end{equation}
\begin{equation}
\begin{aligned}
    \mathcal{L}_{pixel} =& \sum\limits_{h, w}\left(\max_c\left(\frac{1}{2}(\tanh(\frac{b_p}{\gamma}) + 1)\right)\right) \\+&  \sum\limits_{h, w}\left(\max_c\left(\frac{1}{2}(\tanh(\frac{b_n}{\gamma}) + 1)\right)\right). \label{eq:perturb_norm}
\end{aligned}
\end{equation}
In particular, the perturbation is specified in Equation~\ref{eq:perturb}, and the normalization term is represented by $\mathcal{L}_{pixel}$ in Equation~\ref{eq:perturb_norm}. Here, $b_p$ and $b_n$ are the positive and negative components, respectively. The \texttt{clip()} function limits the variable within the range of [0,1]; $h, w, c$ stand for the image's height, width, and channels, respectively; and $\gamma$ is a scaling factor. We refer~\cite{Tao2022BetterTrigger} for comprehensive explanations.

\begin{figure*}[t]
    \centering
    \includegraphics[width=0.85\textwidth]{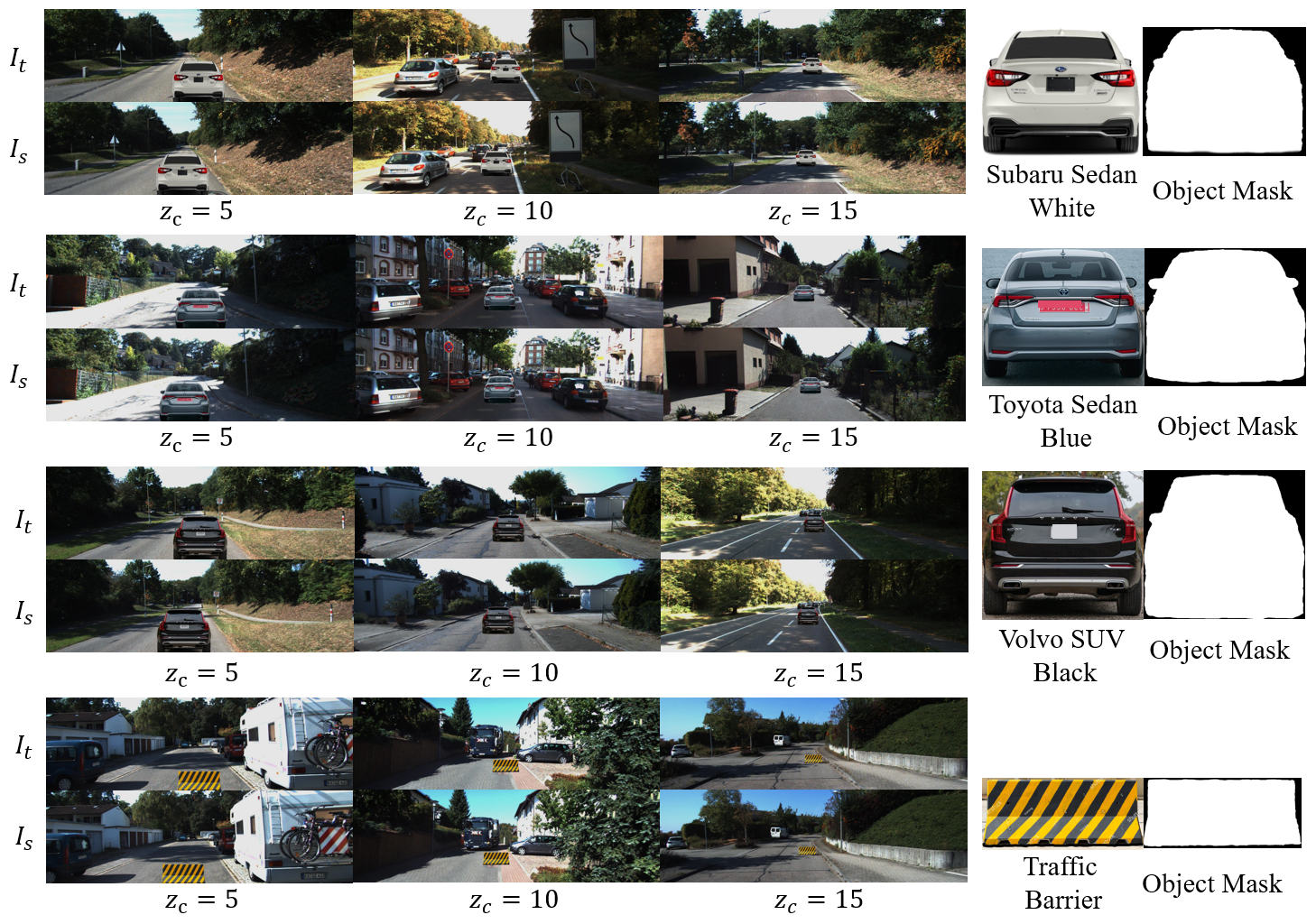}
    \vspace{-10pt}
    \caption{\small More \textbf{examples of view synthesis} with different background scenes, target objects and distance $z_c$ of the object. Object mask is used to remove background of the 2D object image.}
    \vspace{-10pt}
    \label{fig:synthesis-exp}
\end{figure*}

Equation~\ref{eq:perturb_gen} provides the formal representation of our method for generating perturbations. In this equation, $S_{p}$ represents a distribution of physical-world distances and viewing angles (for instance, mirroring the relationships between cameras and vehicles during actual driving scenarios); $S_b$ is the collection of background scenes (such as different street views); $M()$ is the MDE model that predicts the estimated depth; and $MSE()$ is the mean square error.
\begin{gather}\label{eq:perturb_gen}
\scalebox{0.99}{$
    \begin{aligned}
        \min\limits_{b_n, b_p} ~& E_{z_c,\alpha\sim S_p, I^0_t \sim S_b}\left[MSE\left( M\left(I'_t\right)^{-1}, 0 \right)\right] + \mathcal{L}_{pixel}, \\ 
    s.t. ~& L_0(\delta) \leq \epsilon.
    \end{aligned}
$}
\end{gather}
Our adversarial objective is to make the target object appear more distant, so our aim is to maximize the depth estimation (\ie, minimize its reciprocal). Intuitively, we synthesize $I'_t$ with random $I^0_t$ and varied $\alpha$ and $z_c$ of the object $A$ and employ the expectation of transformations (EoT) method ~\cite{athalye2018synthesizing} to enhance physical-world robustness. In the adversarial loss term, we aim to minimize the mean square error between zero and the reciprocal of the depth estimation for the synthesized scenario, while using $\mathcal{L}_{pixel}$ as the normalization term for perturbations. The parameter $\epsilon$ is a predefined $L_0$-norm threshold for perturbations, which signifies the maximum proportion of pixels that can be perturbed (\eg, $\epsilon=1/10$ means at most 1/10 of the pixels within an image can be perturbed for adversarial attack).

\vspace{-10pt}
\section{Evaluation}\label{sec:eval}
In this section, we evaluate the performance of our method in white-box, black-box, and physical-world attack scenarios, and discuss the ablations. Our code is available at: \url{https://github.com/Bob-cheng/DepthModelHardening}
\vspace{-10pt}
\subsection{Experimental Setup}\label{sec:ES}
\textbf{Networks and Dataset.} We$_{\!}$ use$_{\!}$ Monodepth2~\cite{monodepth2}$_{\!}$ and$_{\!}$ DepthHints~\cite{watson2019self}$_{\!}$ as$_{\!}$ our$_{\!}$ subject$_{\!}$ networks$_{\!}$ to$_{\!}$ harden.$_{\!}$ They$_{\!}$ are$_{\!}$ representative$_{\!}$ and$_{\!}$ popular$_{\!}$ self-supervised$_{\!}$ MDE$_{\!}$ models$_{\!}$ that$_{\!}$ are$_{\!}$ widely$_{\!}$ used$_{\!}$ as$_{\!}$ benchmarks$_{\!}$ in$_{\!}$ the$_{\!}$ literature.$_{\!}$ Both$_{\!}$ models$_{\!}$ are$_{\!}$ trained$_{\!}$ on$_{\!}$ the$_{\!}$ KITTI$_{\!}$ dataset~\cite{Geiger2013IJRR}$_{\!}$ and$_{\!}$ our$_{\!}$ methods fine-tune$_{\!}$ the$_{\!}$ original$_{\!}$ models publicly available. 

\smallskip\noindent\textbf{Baselines}\label{app:baseline}
To the best of our knowledge, no previous research has been devoted to harden MDE models, hence there are no immediate benchmarks at our disposal. Therefore, we  opt to expand upon current contrastive learning-based and supervised learning-based adversarial training techniques to MDE, utilizing these as our benchmarks. 

\noindent\textit{(1) Adversarial Contrastive Learning.} Contrastive learning has gained considerable traction as a technique specifically designed for self-supervised learning environments~\cite{chen2020simple,he2020momentum,wu2018unsupervised,tian2020contrastive,ye2019unsupervised,misra2020self}, and it's been employed in combination with adversarial examples for two primary reasons: firstly, to bolster model resilience against adversarial attacks~\cite{kim2020adversarial}, and secondly, to amplify the efficacy of contrastive learning itself~\cite{ho2020contrastive}. In the context of this study, we've taken a leading-edge contrastive learning-based adversarial training technique~\cite{kim2020adversarial} and adapted it to fortify MDE models against physical threats. This adaptation serves as a comparative baseline for our own methodology.

In line with the approach used in previous studies~\cite{kim2020adversarial}, our contrastive learning strategy identifies benign examples (represented as $I_t$) and their corresponding adversarial examples ($I'_t$) as positive pairs, with additional augmentation achieved by modifying color. However, our approach differs in that we do not necessitate negative pairs. Instead, we leverage a learning strategy introduced in SimSiam~\cite{chen2021exploring} that only calls for positive pairs, capable of achieving competitive results with reduced batch sizes.
The fundamental objective is to heighten the similarity between the embeddings of benign and adversarial examples, such that their decoded depth map outputs are analogous. The parameters of the MDE model's encoder and the predictive MLP network are updated in an iterative manner during training.
As we aim to maintain similarity among positive samples, we opt for color augmentation over other transformations like resizing and rotation, as these may influence the depth map output while a change in color wouldn't. The setup of our MLP network and other settings align with those used in SimSiam~\cite{chen2021exploring}, and we direct readers to the SimSiam~\cite{chen2021exploring} study for an in-depth understanding.

\noindent\textit{(2) Supervised Adversarial Learning with Estimated Depth.} In a self-supervised setting, where we do not have access to actual depth ground truth, an alternative method for adversarial training involves using depth estimations generated by the original model with benign sample inputs, serving as pseudo ground truth or pseudo labels, and then carrying out supervised adversarial training. We employ the mean square error (MSE) as the loss function to update the MDE model parameters, aiming to minimize the discrepancy between the model's output for adversarial samples and the pseudo ground truth.

The use of pseudo ground truth, as determined by an existing model, has been shown to be a straightforward and effective strategy within the realm of semi-supervised learning (SSL)~\cite{lee2013pseudo}. It has been applied in adversarial training~\cite{deng2021adversarial} and self-supervised MDE~\cite{petrovai2022exploiting} to enhance model performance. Especially within MDE, the use of pseudo ground truth has been demonstrated to be quite satisfactory when compared to the utilization of actual ground truth. Like our supervised baseline, some studies use depth estimations from an existing MDE model (or pseudo depth labels) to supervise subsequent MDE model training. These investigations have found that a model trained under pseudo-supervision can perform comparably to, or even better than, a model trained using ground truth depth.
In our work, we carry out experiments to compare the performance of a supervised baseline trained with pseudo depth labels and ground truth depth labels. This experimentation substantiates that choosing a pseudo-supervised baseline does not represent a compromise in quality. The detailed results of this comparison can be found in \S\ref{app:sup_GT}.

\begin{table*}[ht]
    \centering
    \caption{\small \textbf{Defensive performance} of original and hardened models under attacks.}
    \vspace{-7pt}
    \label{tab:attack_perf}
    \scalebox{0.95}{
    \begin{threeparttable}
    \begin{tabular}{clcccccccccccccc}
\toprule
 & \multirow{2}{*}{Attacks} & \multicolumn{2}{c}{Original} & \multicolumn{2}{c}{\makecell{L0+SelfSup\\(Ours)}} & \multicolumn{2}{c}{L0+Sup} & \multicolumn{2}{c}{L0+Contras} & \multicolumn{2}{c}{PGD+SelfSup} & \multicolumn{2}{c}{PGD+Sup} & \multicolumn{2}{c}{PGD+Contras} \\ \cline{3-16} 
 &  & \small{ABSE}$\downarrow$ & $\delta\uparrow$ & \small{ABSE}$\downarrow$ & $\delta\uparrow$ & \small{ABSE}$\downarrow$ & $\delta\uparrow$ & \small{ABSE}$\downarrow$ & $\delta\uparrow$ & \small{ABSE}$\downarrow$ & $\delta\uparrow$ & \small{ABSE}$\downarrow$ & $\delta\uparrow$ & \small{ABSE}$\downarrow$ & $\delta\uparrow$ \\ \midrule\midrule
\multirow{8}{*}{\rotatebox[origin=c]{90}{Monodepth2}} 
 & L0 1/20 & 4.71 & 0.65 & \textbf{0.18} & \textbf{0.99} &\underline{ 0.44 }& \underline{0.98} & 1.30 & 0.67 & 0.49 & 0.95 & 0.58 & 0.94 & 0.69 & 0.94 \\
 & L0 1/10 & 6.08 & 0.51 & \textbf{0.25} & \textbf{0.98} & {0.94} & \underline{0.93} & 1.75 & 0.54 & \underline{0.82} & 0.91 & 1.18 & 0.79 & 0.96 & 0.89 \\
 & L0 1/5 & 8.83 & 0.39 & \textbf{0.34 }& \textbf{0.9 }& {1.59} & \underline{0.85} & 2.32 & 0.46 & 2.33 & 0.70 & 2.72 & 0.51 & \underline{1.11} & {0.85} \\
 & L0 1/3 & 9.99 & 0.34 &\textbf{ 0.52} &\textbf{ 0.96} & {2.08} & \underline{0.78} & 2.65 & 0.41 & 4.32 & 0.51 & 4.09 & 0.41 & \underline{1.75} & 0.70 \\
 & PGD 0.05 & 4.74 & 0.56 & \underline{0.82} & \underline{0.97} & 1.29 & 0.80 & 6.61 & 0.38 & \textbf{0.67 }& \textbf{0.98 }& 0.82 & 0.95 & 1.82 & 0.67 \\
 & PGD 0.1 & 11.68 & 0.34 & \underline{1.53} & \underline{0.85} & 2.53 & 0.71 & 12.74 & 0.24 & \textbf{1.38} & \textbf{0.95 }& 1.64 & 0.76 & 2.66 & 0.53 \\
 & PGD 0.2 & 17.10 & 0.23 & \textbf{3.46} & \textbf{0.69} & 6.14 & 0.50 & 20.14 & 0.15 & \underline{3.81} & \underline{0.58} & 5.04 & 0.32 & 3.97 & 0.42 \\
 & Patch & 2.71 & 0.77 & \textbf{0.39} & \textbf{0.98} & 1.35 & 0.89 & 6.40 & 0.52 & \underline{0.40} & \underline{0.98} & 0.84 & 0.92 & 0.50 & 0.95 \\ \hline
\multirow{8}{*}{\rotatebox[origin=c]{90}{DepthHints}}
 & L0 1/20 & 2.33 & 0.66 &\textbf{ 0.19} & \textbf{0.99} & 0.34 & 0.96 & 1.06 & 0.83 & \underline{0.22} & \underline{0.99} & 0.58 & 0.89 & 0.40 & 0.99 \\
 & L0 1/10 & 3.19 & 0.59 &\textbf{ 0.27} & \textbf{0.99} & 0.48 & 0.95 & 1.56 & 0.77 & \underline{0.42} & \underline{0.98} & 1.03 & 0.79 & 0.60 & 0.97 \\
 & L0 1/5 & 4.77 & 0.42 & \textbf{0.40} & \textbf{0.98} & 0.96 & 0.82 & 1.85 & 0.75 & 0.83 & 0.92 & 1.93 & 0.68 & \underline{0.66} & \underline{0.95}  \\
 & L0 1/3 & 6.03 & 0.36 &\textbf{ 0.48} & \textbf{0.98} & 1.64 & 0.68 & 2.60 & 0.69 & 1.45 & 0.79 & 3.06 & 0.57 & \underline{1.16} & \underline{0.82} \\
 & PGD 0.05 & 3.11 & 0.48 & \textbf{0.62 }& \textbf{0.98} & 1.23 & 0.75 & 4.05 & 0.55 & \underline{0.64} & \underline{0.98} & 0.93 & 0.79 & 1.16 & 0.79 \\
 & PGD 0.1 & 6.44 & 0.36 & \underline{1.27} & \underline{0.86} & 2.37 & 0.62 & 7.59 & 0.36 &\textbf{ 1.21 }& \textbf{0.92 }& 1.76 & 0.67 & 1.74 & 0.62  \\
 & PGD 0.2 & 18.37 & 0.23 & \underline{3.09} & \underline{0.60} & 7.13 & 0.41 & 13.59 & 0.24 & 6.14 & \textbf{0.68} & 4.22 & 0.37 & \textbf{2.60} & 0.49 \\
 & Patch & 0.70 & 0.91 & 0.46 & 0.95 & 0.53 & 0.93 & 6.90 & 0.49 & 0.46 & 0.95 & \underline{0.36} & \textbf{0.99} & \textbf{0.34} & \underline{0.98} \\ \bottomrule
\end{tabular}
\begin{tablenotes}
\item *Bold and underlining indicate the best and second-best performance in each row. Hardened models are named the same as in Table 1.
\end{tablenotes}
\end{threeparttable}
}
\vspace{-10pt}
\end{table*}

\begin{figure*}[t]
    \centering
    \includegraphics[width=0.8\textwidth]{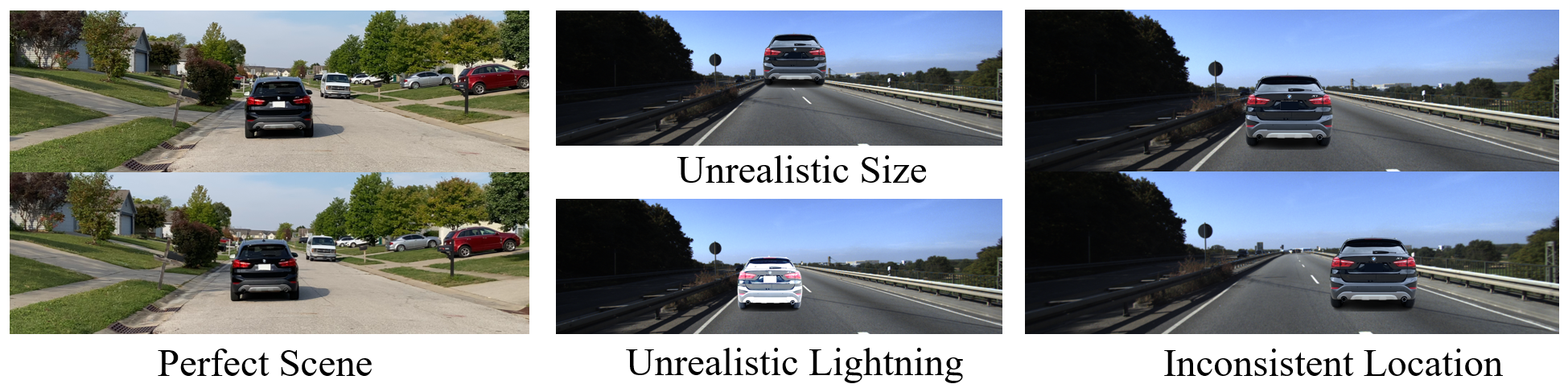}
    \vspace{-5pt}
    \caption{\small The \textbf{reference images} used in our human study.}
    \vspace{-10pt}
    \label{fig:synthesis-exp-ref}
\end{figure*}

\smallskip\noindent\textbf{Training Configurations}\label{app:training_config}
In adversarial training, the ranges of distance $z_c$ and viewing angle $\alpha$ are sampled randomly from 5 to 10 meters and -30 to 30 degrees, respectively. The view synthesis uses EoT~\cite{athalye2018synthesizing}. We generate the adversarial perturbations with two methods:$_{\!}$ $L_0$-norm-bounded$_{\!}$ with$_{\!}$ $\epsilon_{\!}=_{\!}1/10$ and $L_\infty$-norm-bounded ($\ie,$ PGD~\cite{madry2018towards}) with $\epsilon_{\!}=_{\!}0.1$. The latter is for comparison purposes. 
We train with our self-supervised method and two baseline methods based on contrastive learning and supervised learning. Hence, there are 6 approaches combining the 2 perturbation generation methods with the 3 training methods. With these approaches, we fine-tune the original model for 3 epochs on the KITTI dataset and produce 6 hardened models for each network. \added{In our main experiments, we do not intentionally filter out unrealistic paintings like objects on top of others, since those samples that could induce incorrect distance estimation of the target object are infrequent in the entire dataset. Nevertheless, we conduct additional experiments to evaluate the impact on performance in \S\ref{app:inacc_proj}.}

We train our model with one GPU (Nvidia RTX A6000) that has a memory of 48G and the CPU is Intel Xeon Silver 4214R. For each model, doing adversarial training from scratch takes around 70 hours. It includes 20 epochs of training on the KITTI dataset. The fine-tuning of 3 epochs takes about 10 hours. The input resolution of our MDE model is 1024*320 and the original monodepth2 and depthhints models we used for fine-tuning are the official versions trained with both stereo images and videos. In our hardening, we use stereo image pairs with fixed camera pose transformation $T_{t\rightarrow s}$. In perturbation generation, we use 10 steps and a step size of $2.5 \cdot\epsilon / 10$ in $L_2$ and $L_{\infty}$-bounded attacks to ensure that we can reach the boundary of the
$\epsilon$-ball from any starting point within it \cite{madry2018towards} and a batch size of 12. In MDE training, the batch size is 32, and the learning rate is 1e-5. We use Adam as the optimizer and other training setups are the same as the original model. 

As for the selection of 2D images of objects, as shown in Figure~\ref{fig:synth_def} (a) and Figure~\ref{fig:synth_def} (b), we have assumptions about the initial relative positions between the target object and the camera (\ie, the 3D coordinates of the center of the object is $(0, 0, z_c)$ in the camera's coordinate system and the viewing angle $\alpha$ of the camera is 0 degree). Hence, for a more realistic and high-quality synthesis, the camera should look at the center of the target object at the same height while taking the 2D image of the object. The width $w$ and height $h$ of the 2D image of the object should be proportional to the physical size $W$ and $H$ of it: 
$w/W = h/H.$ Moreover, when we prepared the 2D image of the object, we also prepared a corresponding mask to ``cut out” the main body of the object for projection and we take the object together with its shadow to preserve reality. Examples of object masks can be found in Figure~\ref{fig:synthesis-exp}.

We train models with $L_0$ and $L_{\infty}$-bounded (\ie, PGD) perturbations in our evaluation but not $L_2$ norm because \cite{madry2018towards} has demonstrated that models hardened with $L_\infty$-bounded perturbations are also robust against $L_2$-bounded attacks and our experiments in \S\ref{app:more_atks} also validate the robustness of our models. In addition, physical-world attacks with adversarial patches have more resemblance to $L_0$-bounded attacks that only restrict the ratio of perturbed pixels rather than the magnitude of the perturbation.

\smallskip\noindent\textbf{Attacks.} We conduct various kinds of attacks to evaluate the robustness of different models. They are $L_0$-norm-bounded attacks with $\epsilon=1/20, 1/10, 1/5 \text{ and } 1/3$, $L_\infty$-norm-bounded (PGD) attacks with $\epsilon=0.05, 0.1 \text{ and } 0.2$ (image data are normalized to [0,1]), and an adversarial patch attack in~\cite{mathew2020monocular}. Adversarial perturbation or patch is applied to an object image. The patch covers 1/10 of the object at the center. Each attack is evaluated with 100 randomly selected background scenes. The object is placed at a distance range of 5 to 30 meters and a viewing angle range of -30 to 30 degrees. We report the average attack performance over different background scenes, distances, and viewing angles for each attack. In addition, we conduct the SOTA physical-world attack~\cite{cheng2022physical} with the printed optimal patch and a real vehicle in driving scenarios. Adversarial examples are in 
\S\ref{append:adv_egs}. Evaluation with more attacks are in \S\ref{app:more_atks}.

\smallskip\noindent\textbf{Evaluation Metrics}\label{app:eval_metrics}
In particular, the evaluation metrics we used in our evaluation are defined as follows, where we use $X=\{x_1, x_2, ..., x_n\}$ to denote the estimated depth map and $Y=\{y_1, y_2, ..., y_n\}$ to denote the reference depth map and $I()$ is the indicator function that evaluates to 1 only when the condition is satisfied and 0 otherwise.
\begin{gather}
\scalebox{0.9}{$
\begin{aligned}
    ABSE &= \frac{1}{n}\sum\limits_{i=1}^{n}|x_i - y_i|,\quad RMSE = \sqrt{\frac{1}{n}\sum\limits_{i=1}^{n}(x_i - y_i)^2}\\
    ABSR &= \frac{1}{n}\sum\limits_{i=1}^{n}(\frac{|y_i - x_i|}{y_i} ), \quad SQR = \frac{1}{n}\sum\limits_{i=1}^{n}\frac{(y_i - x_i)^2}{y_i} \\
    \delta &= \frac{1}{n}\sum\limits_{i=1}^{n} I(\max\{\frac{x_i}{y_i}, \frac{y_i}{x_i}\} < 1.25)
\end{aligned}
$}
\end{gather}
The mean absolute error (ABSE) and root mean square error (RMSE) are common metrics and are easy to understand. Intuitively, the relative absolute error (ABSR) is the mean ratio between the error and the ground truth value, and the relative square error (SQR) is the mean ratio between the square of error and the ground truth value. $\delta$  denotes the percentage of pixels of which the ratio between the estimated depth and ground-truth depth$_{\!}$ is$_{\!}$ smaller$_{\!}$ than$_{\!}$ 1.25.

\begin{figure*}[t]
  \centering
\includegraphics[width=0.85\textwidth]{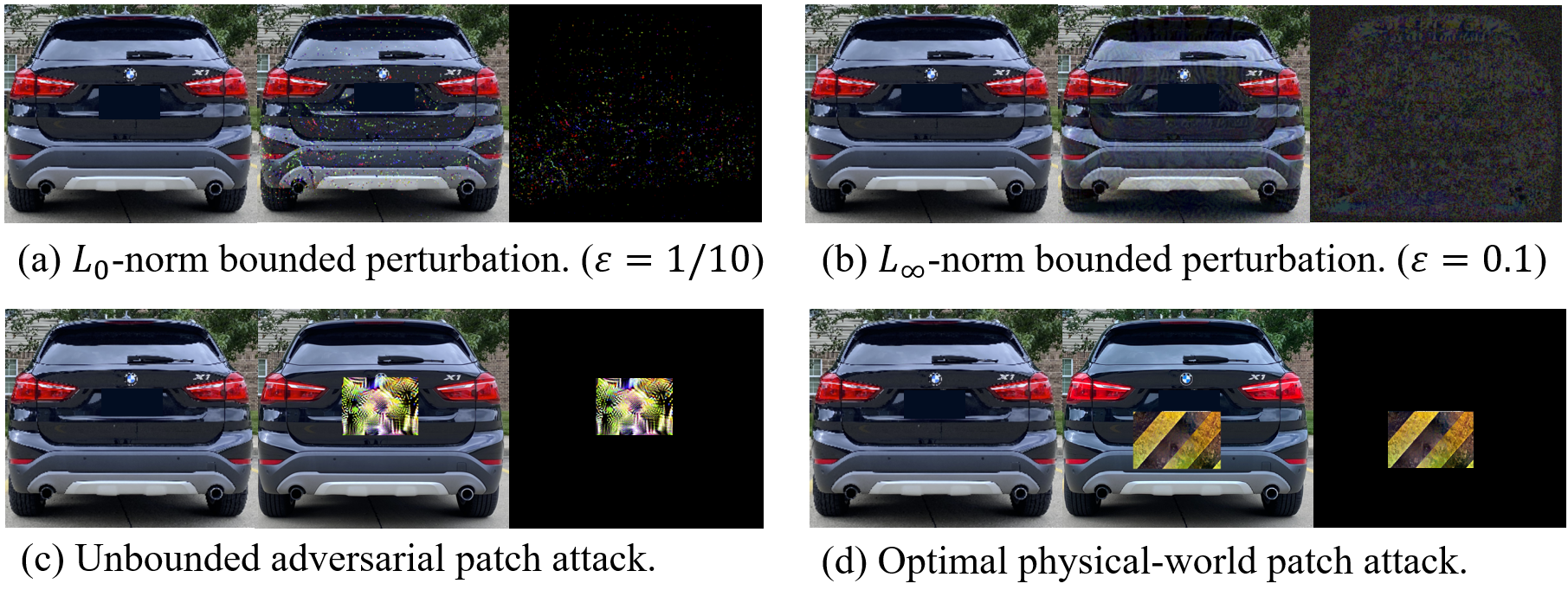}
\vspace{-5pt}
\caption{ \small \textbf{Examples of adversarial attacks} in our robustness evaluation.}
\label{fig:adv_examples}
\vspace{-10pt}
\end{figure*}

\begin{table}[t]
    \centering
    \caption{\small Human evaluations of the \textbf{quality of our synthesized images}. We show the number of participants who gave a score in the corresponding range in each row.}
    \label{tab:syn-quality-eval}
    \vspace{-5pt}
    \scalebox{0.9}{
    \begin{tabular}{ccccc}
    \toprule
\textbf{Score Ranges} & \textbf{Size }& \textbf{Location} & \textbf{Lightning} & \textbf{Overall} \\\midrule\midrule
\textbf{1-2} & 0 & 0 & 1 & 0 \\ 
\textbf{5-6} & 4 & 6 & 6 & 3 \\ 
\textbf{7-8} & 16 & 24 & 23 & 18 \\ 
\textbf{9-10} & 42 & 48 & 40 & 45 \\ 
\textbf{Total} & 100 & 100 & 100 & 100\\ \midrule
\textbf{Average Score} & 7.7 & 7.21 & 7.43 & 7.74\\
\bottomrule
\end{tabular}
}
\vspace{-10pt}
\end{table}

\vspace{-5pt}
\subsection{Main Results}\label{sec:main_results}

\subsubsection{Benign Performance}  Together with the original model, we have 7 models under test for each network. We evaluate the depth estimation performance on the KITTI dataset using the Eigen split and report the results in Table~\ref{tab:ben_perf}. 
As shown, self-supervised and supervised methods have little influence on the models' depth estimation performance, which means these approaches can harden the model with nearly no benign performance degradation. In contrast, the contrastive learning-based approach performs the worst. The ABSE of estimated depth is over 1 m worse than the original model. The reason could be that contrastive learning itself does not consider the specific task ($\ie,$ MDE) but fine-tunes the encoder to filter out the adversarial perturbations. Thus the benign performance is sacrificed during training. The other two methods consider the depth estimation performance either by preserving the geometric relationship of 3D space in synthesized frames or by supervising the training with estimated depth.

\begin{figure*}[t]
    \centering
    \includegraphics[width=0.8\textwidth]{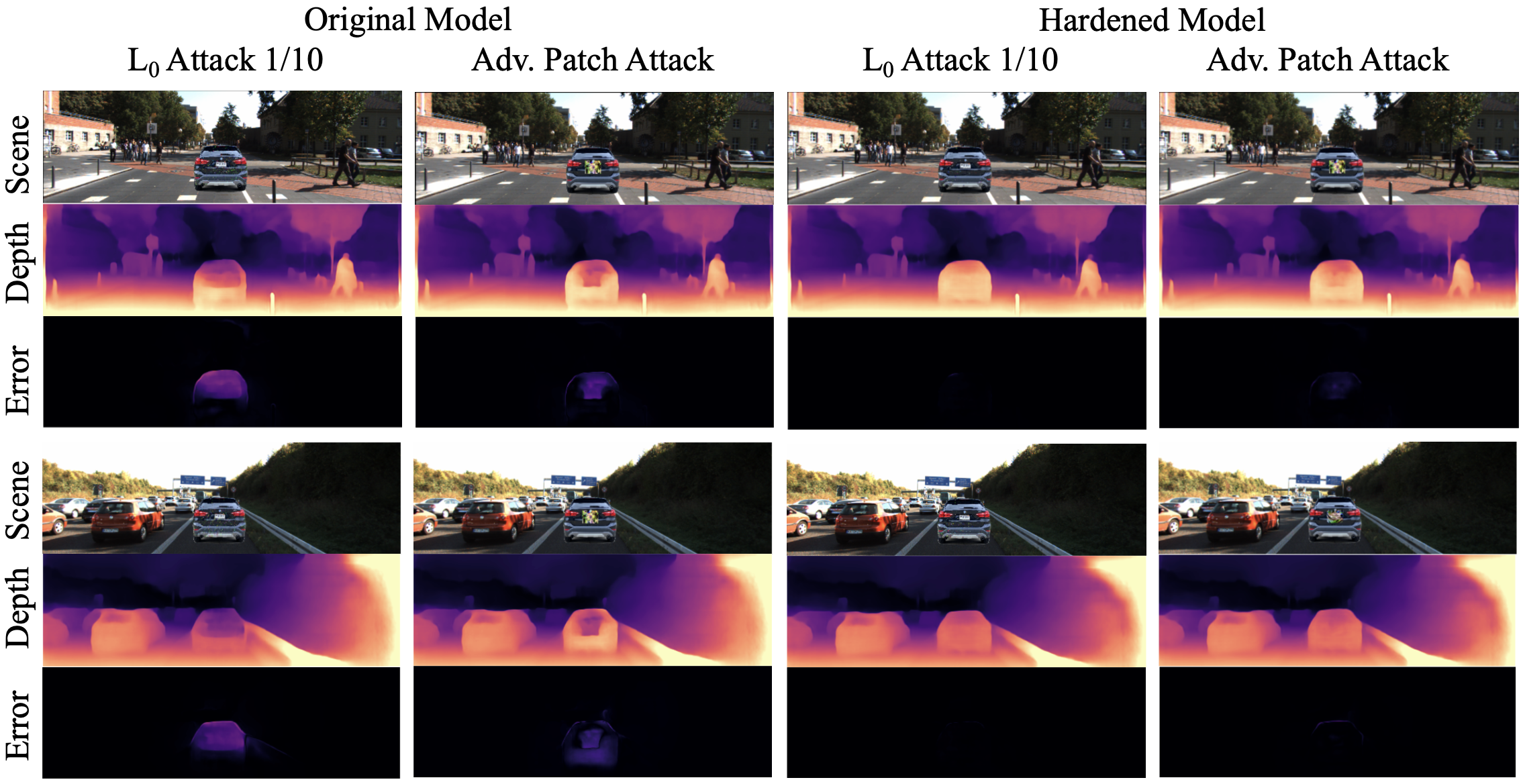}
    \includegraphics[width=0.8\textwidth]{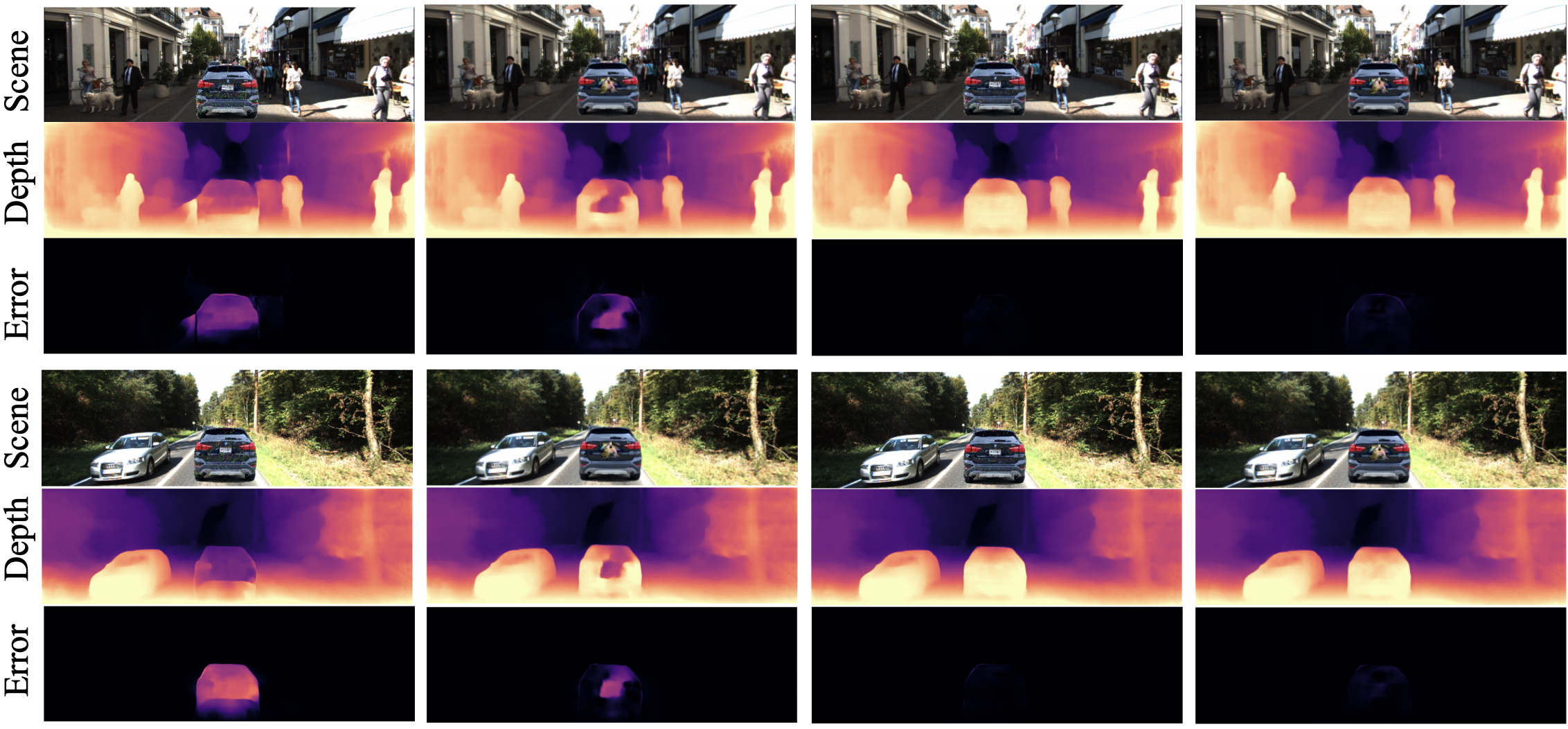}
    \vspace{-5pt}
    \caption{\small \textbf{Qualitative results} of the defensive performance of our hardened model. }
    \label{fig:qual_exps}
    \vspace{-10pt}
\end{figure*}

\vspace{-7pt}
\subsubsection{Quality of the view synthesis}\label{app:syn-quality}

To demonstrate the quality of our view synthesis, we show more examples of the synthesized images in Figure~\ref{fig:synthesis-exp}. $I_s$ and $I_t$ are two adjacent views of the same scene. Different target objects (e.g., the white sedan, black SUV and traffic barrier) are synthesized into the views with various distances $z_c$ and background scenes. To evaluate the veracity of these images, we use Amazon Mechanical Turk to conduct human evaluations of our synthesized images. Participants are requested to evaluate the quality of the synthesized objects from 4 distinct perspectives: size, consistency of location, lighting, and overall quality. The score ranges from 1 to 10 for each metric. We use a score of 1 to indicate scenes synthesized with random projections and  a score of 10  for real scenes taken by stereo cameras. Examples of perfect (score of 10) and unrealistic  (score of 1) scenes for each metric are provided as references (See Figure~\ref{fig:synthesis-exp-ref}). The results of the experiment with 100 participants can be found in Table~\ref{tab:syn-quality-eval}. In each row, we show the number of participants who gave a score within the corresponding range and summarized the average in the last row. As demonstrated, We got an average score of 7.7 regarding the size, 7.21 regarding the location,  7.43 regarding the lighting, and 7.74 regarding the overall quality. Compared with the real scene, our synthesized scene is slightly inferior but still much better than random projection. More importantly, it$_{\!}$ works$_{\!}$ well$_{\!}$ to$_{\!}$ harden$_{\!}$ models$_{\!}$ against$_{\!}$ real$_{\!}$ attacks$_{\!}$ at$_{\!}$ a$_{\!}$ low$_{\!}$ cost.

\begin{table}[t]
    \centering
    \caption{\small Defensive performance of original and hardened models under\textbf{ black-box attacks}. }
    \label{tab:black_box}
    \vspace{-5pt}
    \scalebox{0.87}{
    \begin{threeparttable}
    \begin{tabular}{lcccccccc}
\toprule
\multicolumn{1}{r}{Target} & \multicolumn{2}{c}{Original} & \multicolumn{2}{c}{\makecell{L0+SelfSup\\(Ours)}} & \multicolumn{2}{c}{L0+Sup} & \multicolumn{2}{c}{L0+Contras} \\ \cline{2-9} 
Source & ABSE & $\delta\uparrow$ &ABSE & $\delta\uparrow$ & ABSE & $\delta\uparrow$& ABSE & $\delta\uparrow$\\ \midrule\midrule
Original & - & - & \textbf{0.25} & \textbf{0.99} & 0.55 & 0.95 & 1.35 & 0.71 \\
L0+SelfSup & \textbf{0.52} & \textbf{0.93} & - & - & \textbf{0.24} & \textbf{0.98} & \textbf{0.45} & \textbf{0.95} \\
L0+Sup & 1.09 & 0.80 & \textbf{0.29} & \textbf{0.99} & - & - & 0.65 & 0.89 \\
L0+Contras & 2.65 & 0.50 & \textbf{0.22} & \textbf{0.99} & 0.27 & 0.98 & - & - \\ \bottomrule
\end{tabular}
\begin{tablenotes}
\item *Bold indicates the best performance in each row or column.
\end{tablenotes}
\end{threeparttable}
}
\vspace{-10pt}
\end{table}

\begin{figure*}[t]
\centering
\begin{minipage}{.23\textwidth}
 \centering
    \includegraphics[width=\textwidth]{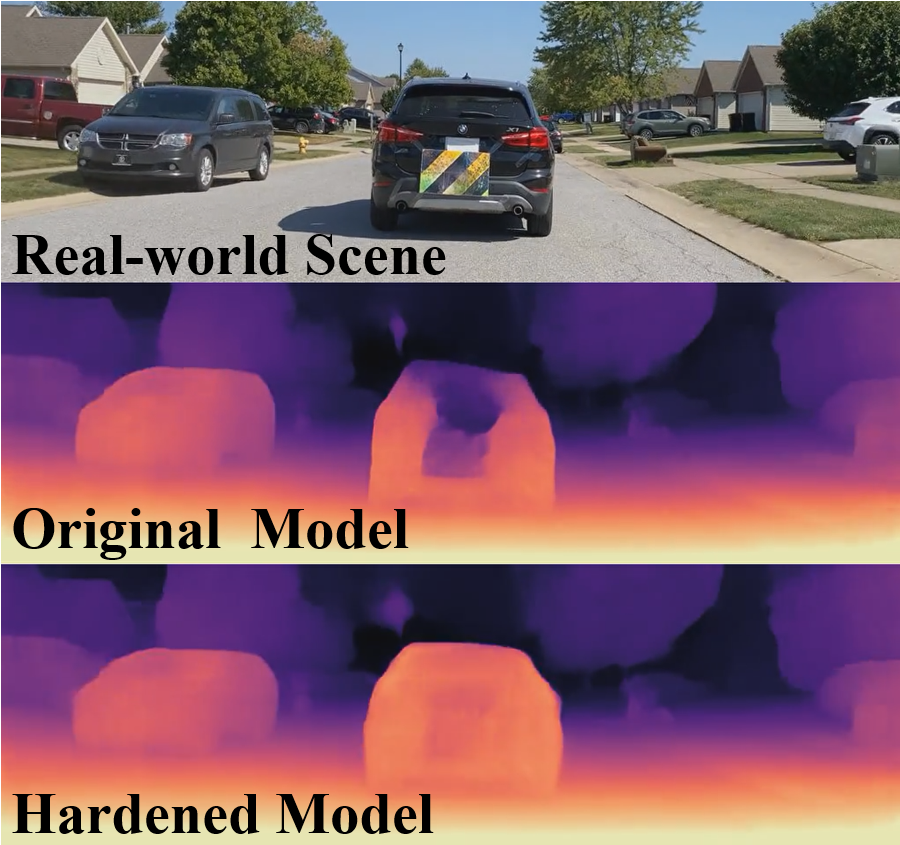}
    \caption{\small \textbf{Physical-world} attack and defense. Video: {\footnotesize \url{https://youtu.be/_b7E4yUFB-g}}}
    \label{fig:physical_atk}
\end{minipage}%
\begin{minipage}{.6\textwidth}
  \centering
    \begin{subfigure}[ht]{0.44\columnwidth}
        \centering
        \includegraphics[width=\columnwidth]{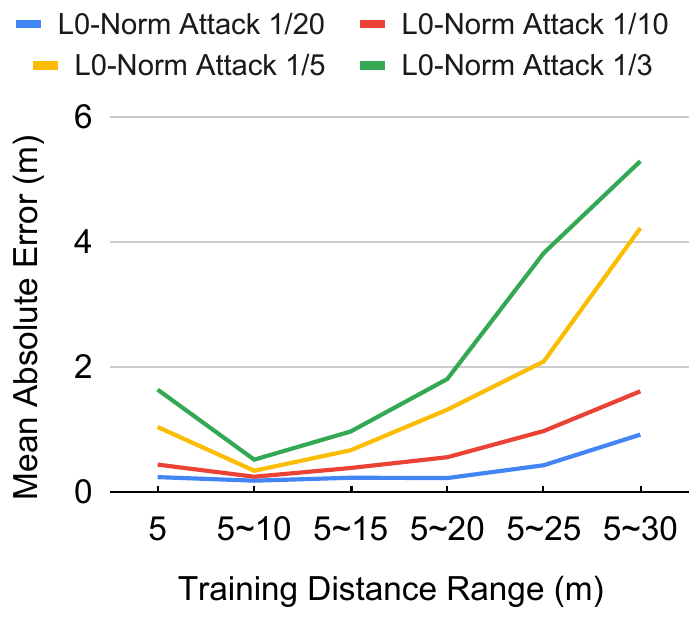}
        \caption{\small $L_0$-norm attack. }
        \label{fig:attack_dist_l0}
    \end{subfigure}
    \hspace{5pt}
    \begin{subfigure}[ht]{0.45\columnwidth}
        \centering
        \includegraphics[width=\columnwidth]{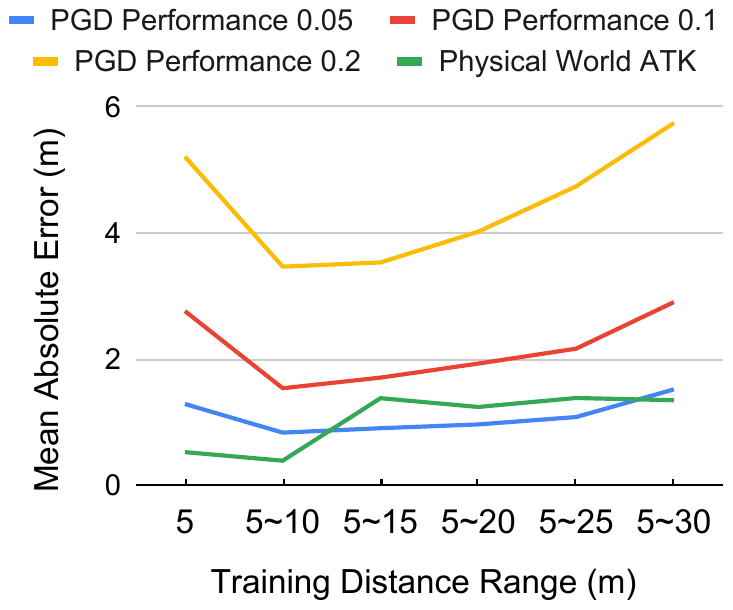}
        \caption{\small PGD and patch attack. }
        \label{fig:attack_dist_pgd}
    \end{subfigure}   
\vspace{-5pt}
\caption{\small Defensive performance with different \textbf{training distance ranges}.}\label{fig:attack_dist}
\end{minipage}
\vspace{-10pt}
\end{figure*}

\begin{figure*}[t]
\centering
\includegraphics[width=0.85\textwidth]{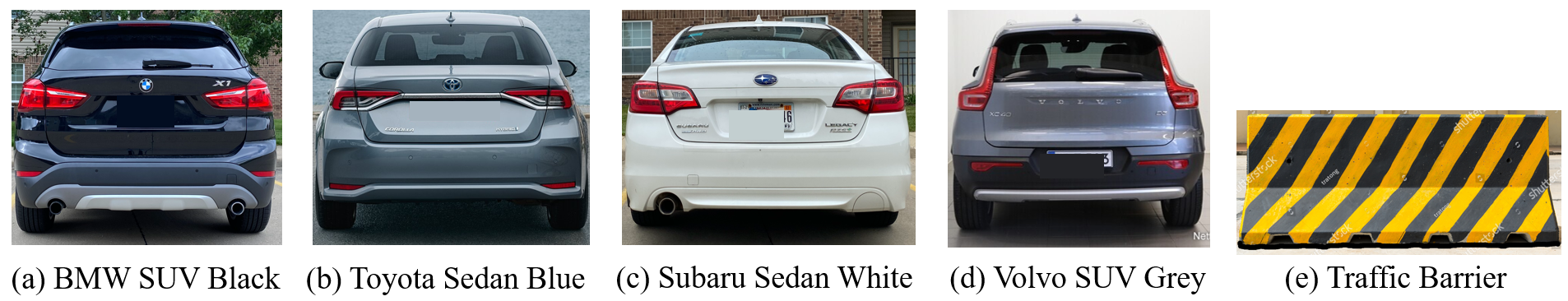}
\vspace{-5pt}
\caption{ \small The\textbf{ various target objects} in the transferability evaluation.}
\label{fig:target_objects}
\vspace{-10pt}
\end{figure*}

\vspace{-7pt}
\subsubsection{White-box Attacks} We conduct various white-box attacks on each model to evaluate the robustness of hardened models. Specifically, for each model, we compare the estimated depth of the adversarial scene ($\ie,$ $I_{t}'$) with that of the corresponding benign scene ($\ie,$ $I_{t}$ in Equation~\ref{eq:it_syn}) and larger difference means worse defensive performance. Table~\ref{tab:attack_perf} shows the result. As shown, all the hardened models have better robustness than the original models against all kinds of attacks, and it is generic on the two representative MED networks, which validates the effectiveness of our adversarial training approaches. Comparing different approaches, \texttt{L0+SelfSup} has the best performance in general. It reduces the ABSE caused by all-level $L_0$-norm-bounded attacks from over 4.7~m to less than 0.6~m. Specifically,$_{\!}$ the$_{\!}$ self-supervision-based$_{\!}$ method$_{\!}$ outperforms$_{\!}$ the$_{\!}$ contrastive$_{\!}$ learning-based$_{\!}$ and$_{\!}$ the$_{\!}$ supervision-based$_{\!}$ methods$_{\!}$ regardless$_{\!}$ of$_{\!}$ the$_{\!}$ perturbation$_{\!}$ generation$_{\!}$ method$_{\!}$ used. It is because the self-supervision-based method follows the original training procedure that is carefully designed for the network and has been evaluated thoroughly. It is not$_{\!}$ surprising$_{\!}$ that$_{\!}$ models$_{\!}$ adversarially$_{\!}$ trained$_{\!}$ with$_{\!}$ $L_0$-norm-bounded$_{\!}$ perturbation$_{\!}$ (our$_{\!}$ method)$_{\!}$ achieve better robustness against $L_0$-norm-bounded attacks and so do PGD-based methods, but more importantly, $L_0$-norm-based training also has good defensive performance against PGD attacks. The robustness of \texttt{L0+SelfSup} is only slightly worse than \texttt{PGD+SelfSup} on some PGD attacks and even better than it on stronger PGD attacks. An explanation is that $L_0$-norm does not restrict the magnitude of perturbation on each pixel, and stronger PGD attacks are closer to this setting ($\ie,$ high-magnitude perturbations) and can be well-defended using the $L_0$-norm-based adversarial training. Monodepth2 is vulnerable to the patch attack, and this kind of attack can be well-defended by our methods. \texttt{L0+SelfSup} also performs the best. Depthhints itself is relatively robust to the patch attack, and our methods can further reduce the attack effect. Our defense generalizes well to complex scenes including  various road types, driving speed, and the density of surrounding objects. Qualitative results are in \S\ref{append:adv_egs}.

\begin{table}[t]
    \centering
    \caption{\small Ablations on \textbf{the range of angles} in adversarial training with \texttt{L0+SelfSup}.}\label{tab:abl_angle}
    \vspace{-5pt}
    \scalebox{0.95}{
    \begin{tabular}{cccccc}
\toprule
        & $0^\circ$     & $10^\circ$ & $ 20^\circ$   & $ 30^\circ$   & $ 40^\circ$   \\ 
  Attacks     & ABSE$\downarrow$ & ABSE$\downarrow$ & ABSE$\downarrow$ & ABSE$\downarrow$ & ABSE$\downarrow$\\\midrule \midrule
L0 1/10 & 0.271 & 0.260        & 0.253          & 0.251          & \textbf{0.249} \\ 
L0 1/5  & 0.363 & 0.352        & 0.351          & \textbf{0.347} & 0.350          \\ 
PGD 0.1 & 1.662 & 1.543        & 1.544          & 1.539          & \textbf{1.536} \\
PGD 0.2 & 3.587 & 3.472        & \textbf{3.465} & 3.467          & 3.470          \\ \bottomrule
\end{tabular}
}
\vspace{-10pt}
\end{table}

\begin{table}[t]
    \centering
    \caption{\small Defensive performance on \textbf{various target objects}.
    }
    \label{tab:target_obj}
    \vspace{-5pt}
    \scalebox{0.84}{
    \begin{threeparttable}
    \begin{tabular}{lcccccccc}
\toprule
\multicolumn{1}{r}{Models} & \multicolumn{2}{c}{Original} & \multicolumn{2}{c}{\makecell{L0+SelfSup\\(Ours)}} & \multicolumn{2}{c}{L0+Sup} & \multicolumn{2}{c}{L0+Contras} \\ \cline{2-9} 
Objects & ABSE & $\delta\uparrow$ &ABSE & $\delta\uparrow$ & ABSE& $\delta\uparrow$& ABSE& $\delta\uparrow$\\ \midrule\midrule
SUV Black& 6.079 & 0.512 & \textbf{0.248}& \textbf{0.988} &\underline{0.949} & \underline{0.934} &1.642  & 0.552 \\
SD Blue  & 4.259 & 0.456 & \textbf{0.651} & \textbf{0.905} & \underline{1.097} & \underline{0.833} & 1.701  & 0.606 \\
SD White & 4.262 & 0.437 & \textbf{1.408}& \textbf{0.776} & 2.602& \underline{0.632} &\underline{2.324}  &0.505  \\
SUV Grey &6.379 & 0.535 &\textbf{1.565} & \textbf{0.889} & \underline{1.859}& \underline{0.836} & 2.476 &0.552  \\
Barrier &4.479 & 0.375 & \underline{1.619}& \underline{0.54} &2.3 &0.483  &\textbf{0.813}  & \textbf{0.767} \\
\bottomrule
\end{tabular}
\begin{tablenotes}
\item *Abbreviations. SD: Sedan.
\end{tablenotes}
\end{threeparttable}
}
\vspace{-15pt}
\end{table}

\vspace{-7pt}
\subsubsection{Black-box Attacks} 
We$_{\!}$ also$_{\!}$ validate$_{\!}$ our$_{\!}$ methods$_{\!}$ against$_{\!}$ black-box$_{\!}$ attacks.$_{\!}$ We$_{\!}$ use$_{\!}$ the$_{\!}$ original$_{\!}$ Monodepth2 model$_{\!}$ and the models fine-tuned with $L_0$-norm-bounded perturbations and the three training methods. We perform $L_0$-norm-bounded attacks on each model with$_{\!}$ $\epsilon=1/10$$_{\!}$ and apply the generated adversarial object to other models evaluating the black-box attack performance.$_{\!}$ The$_{\!}$ first$_{\!}$ column$_{\!}$ in$_{\!}$ Table~\ref{tab:black_box}$_{\!}$ denotes$_{\!}$ the$_{\!}$ source$_{\!}$ models$_{\!}$ and$_{\!}$ other$_{\!}$ columns$_{\!}$ are the target models' defensive performance.$_{\!}$ Looking at each column,  adversarial examples generated from \texttt{L0+SelfSup} have the worst attack performance, which indicates the low transferability of adversarial examples generated from the model trained with our self-supervised methods. From$_{\!}$ each$_{\!}$ row,$_{\!}$ we$_{\!}$ can$_{\!}$ observe$_{\!}$ that$_{\!}$ \texttt{L0+SelfSup}$_{\!}$ has$_{\!}$ the$_{\!}$ best$_{\!}$ defensive$_{\!}$ performance$_{\!}$ against$_{\!}$ adversarial$_{\!}$ examples$_{\!}$ generated$_{\!}$ from$_{\!}$ each$_{\!}$ source$_{\!}$ model,$_{\!}$ which$_{\!}$ further$_{\!}$ validates$_{\!}$ the$_{\!}$ robustness$_{\!}$ of$_{\!}$ \texttt{L0+SelfSup}.$_{\!}$ In$_{\!}$ summary,$_{\!}$ the$_{\!}$ self-supervised$_{\!}$ training$_{\!}$ method$_{\!}$ can$_{\!}$ produce$_{\!}$ safer$_{\!}$ and$_{\!}$ more$_{\!}$ robust$_{\!}$ models.$_{\!}$ Their$_{\!}$ examples$_{\!}$ have$_{\!}$ low$_{\!}$ transferability,$_{\!\!}$ and\!$_{\!}$ they$_{\!}$ defend$_{\!}$ against$_{\!}$ black-box$_{\!}$ attacks$_{\!}$ well.

\vspace{-7pt}
\subsubsection{Physical-world Attacks}  Our evaluation with the state-of-the-art physical-world MDE attack~\cite{cheng2022physical} validates the effectiveness of our method in various real-world lighting conditions, driving operations, road types, etc. The experimental settings are the same as~\cite{cheng2022physical}. Figure~\ref{fig:physical_atk} shows the result. The first row is the real-world adversarial scene, in which a car is moving with an adversarial patch attached to the rear. The second row is the depth predicted by the original Monodepth2 model and the third row is predicted by our hardened model (\texttt{L0+SelfSup}). The ``hole'' in the output of the original model caused by the adversarial patch is fixed in our hardened model. It is known that the adversarial attacks designed for the physical world are still generated in the digital world and they have better digital-world attack performance than physical-world performance because additional environmental variations in the real world degrade the effectiveness of adversarial patterns~\cite{braunegg2020apricot,wu2020making,cheng2022physical}. Hence, defending attacks in the digital world is more difficult and our success in digital-world defense in previous experiments has implied effectiveness in the real world.

\vspace{-7pt}
\subsubsection{Qualitative Results against Adversarial Attack} \label{append:adv_egs}

Figure~\ref{fig:adv_examples} gives examples of the three kinds of adversarial attacks we conducted in our robustness evaluation. The first column is the original object; the second column is the adversarial one and the third column is the adversarial perturbations. We scale the adversarial perturbations of the $L_{\infty}$-norm-bounded attack for better visualization. $L_0$-norm restricts the number of perturbed pixels. $L_{\infty}$-norm restricts the magnitude of perturbation of each pixel. Adversarial patch attack perturbs pixels within the patch area.

\begin{figure}[t]
    \centering
    \includegraphics[width=0.65\columnwidth]{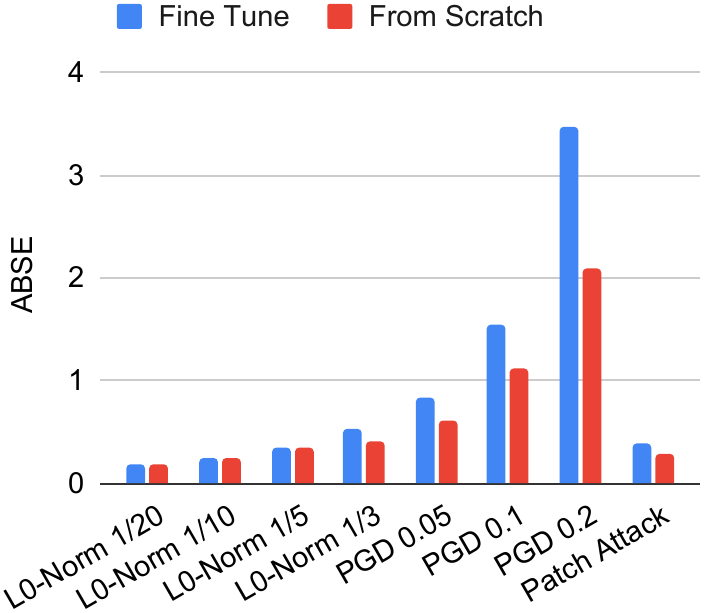}
    \vspace{-2mm}
    \caption{ \small Performance of \textbf{fine-tuning and training from scratch}.}\label{fig:finetune}
    \vspace{-10pt}
\end{figure}

Figure~\ref{fig:qual_exps} shows the qualitative results of our evaluation. The vertical axis denotes different street views and the horizontal axis denotes different models and attacks. Here we compare the performance of the original Monodepth2 model with our model hardened by \textbf{L0+SelfSup}. For each street view, model, and attack, the first row is the adversarial scene (\ie, the scene with the adversarial object), the second row is the corresponding depth estimation and the third row is the depth estimation error caused by the adversarial perturbations. As shown, our method mitigates the effectiveness of attacks significantly and the attacks can hardly cause any adversarial depth estimation error on our hardened model. In addition, our method works on different scenes including complex ones that have multiple pedestrians or vehicles at different speeds. It is because we do not have assumptions about the scene geometry or surrounding objects in our method. Both the scene synthesis and depth estimation are single-image-based and the scene geometry will not affect the quality of synthesis.

\begin{table*}[t]
    \centering
    \caption{\small Defensive performance of original and hardened models under \textbf{more attacks}.}
    \label{tab:perf_more_atks}
    \vspace{-5pt}
    \scalebox{1}{
    \begin{tabular}{lcccccccc}
\toprule
\multirow{2}{*}{Attacks} & \multicolumn{2}{c}{Original} & \multicolumn{2}{c}{L0+SelfSup (Ours)} & \multicolumn{2}{c}{L0+Sup} & \multicolumn{2}{c}{L0+Contras} \\ \cline{2-9} 
 & ABSE$\downarrow$ & $\delta \uparrow$ & ABSE$\downarrow$ & $\delta \uparrow$ & ABSE$\downarrow$ & $\delta \uparrow$ & ABSE$\downarrow$ & $\delta \uparrow$ \\ \midrule\midrule
$L_2$-PGD $\epsilon$ = 8 & 1.403 & 0.76 & \textbf{0.294} & \textbf{0.996} & 1.161 & 0.741 & 0.66 & 0.919 \\
$L_2$-PGD $\epsilon$ = 16 & 6.491 & 0.522 & \textbf{0.597} & \textbf{0.984} & 2.516 & 0.479 & 1.437 & 0.734 \\
$L_2$-PGD $\epsilon$ = 24 & 13.018 & 0.354 & \textbf{0.932} & \textbf{0.913} & 3.613 & 0.387 & 2.92 & 0.7 \\
APGD $\epsilon$ = 0.05 & 5.557 & 0.739 & \textbf{0.423} & \textbf{0.976} & 1.614 & 0.793 & 2.71 & 0.859 \\
APGD $\epsilon$ = 0.1 & 10.216 & 0.46 & \textbf{0.928} & \textbf{0.945} & 3.279 & 0.603 & 4.578 & 0.793 \\
\begin{tabular}[l]{@{}l@{}}Square Attack \\$\epsilon$ = 0.1 N=5000\end{tabular} & 0.924 & 0.924 & \textbf{0.422} & \textbf{0.991} & 0.712 & 0.934 & 0.568 & 0.973 \\ 
Gaussian Blur & 0.323 & 0.996& \textbf{0.191} & \textbf{0.997} & 0.288 &0.997 & 0.264 &0.997 \\
AdvLight&  0.512 & 0.988  & \textbf{ 0.493} & \textbf{0.991 } &  0.513 & 0.987 & 0.504  & 0.988  \\
\bottomrule
\end{tabular}
}
\vspace{-10pt}
\end{table*}

\vspace{-10pt}
\subsection{Ablations}

\subsubsection{Distance Range in Training.} While synthesizing views in training, the range of distance $z_c$ of the target object is randomly sampled from $d_1$ to $d_2$ meters. In this ablation study, we evaluate the effect of using different ranges of distance in training. We use  \texttt{L0+SelfSup} to fine-tune the original Monodepth2 model. 
The ranges of distance we use in training are [5, 5], [5, 10], [5, 15], [5, 20], [5, 25] and [5, 30] (Unit: meter). 
Note that, for a fair comparison, the range of distance we use in model evaluation is always from 5 to 30 meters. The results are shown in Figure~\ref{fig:attack_dist}. As shown, the model trained with a distance range of 5-10 meters has the best robustness and a larger or smaller distance range could lead to worse performance. It is because further distances lead to smaller objects on the image and fewer pixels are affected by the attack. Thus the attack performance is worse at further distances and training with these adversarial examples is not the most effective. If the distance in training is too small (\eg 5 meters), the model cannot defend various scales of attack patterns and cannot generalize well to further distances. In our experiments, the range of 5-10 meters makes a good balance between training effectiveness and generality. Other ablation studies about viewing angles range in training are in \S\ref{app:ang_range}, transferability to unseen target objects is in \S\ref{append:target_objs}, comparing training from scratch and fine-tuning is in 
\S\ref{app:scratch}
and the performance of method combinations can be found in 
\S\ref{app:combi}.

\begin{table}[t]
\centering
\caption{\small \textbf{Benign performance} of models trained with \textbf{methods combinations}.}
\vspace{-5pt}
\label{tab:comb_benign}
\scalebox{1}{
\begin{tabular}{lccccc}
\toprule
& ABSE$\downarrow$ & RMSE$\downarrow$ & ABSR$\downarrow$ & SQR$\downarrow$ & $\delta\uparrow$  \\ \midrule\midrule
Original & 2.125 & 4.631 & 0.106 & 0.807 & 0.877 \\
SelfSup+Con & 2.172 & 4.84 & 0.105 & 0.847 & 0.875  \\
SelfSup+Sup & 2.161 & 4.819 & 0.105 & 0.836 & 0.875   \\
Con+Sup & 2.408 & 4.986 & 0.122 & 0.93 & 0.849  \\
All & 2.171 & 4.83 & 0.105 & 0.841 & 0.875  \\
\bottomrule
\end{tabular}
}
\vspace{-10pt}
\end{table}

\vspace{-7pt}
\subsubsection{Viewing Angles Range in Training}\label{app:ang_range}

Randomizing the degree of synthesis will make the model more robust in the physical-world settings because the camera's viewing angle is not fixed toward the target object in practice, and the range we use (-30 degrees to 30 degrees) covers the most common situations. Ablations of the range of angles during adversarial training do not have as much effect as changing the range of distance because distance has a dominant influence on the size (\ie, number of pixels) of the adversarial object on the synthesized scene image (\ie, the further object looks smaller) and the number of adversarial pixels affects the attack performance a lot~\cite{brown2017adversarial}. We conduct experiments with different viewing angle ranges, and other settings are the same as the ablation study of distance ranges in Training. Table~\ref{tab:abl_angle} shows the result. As shown, ranges of viewing angles have less influence on the defensive performance, and using a fixed setting (\ie, $0^\circ$) still performs the worst, which is consistent with our claims.

\begin{table*}[t]
    \centering
    \caption{\small \textbf{Defensive performance} of models trained with \textbf{methods combinations}.}
    \vspace{-5pt}
    \label{tab:comb_perf}
    \scalebox{1}{
    \begin{threeparttable}
    
    \begin{tabular}{clcccccccccc}
\hline
 & \multirow{2}{*}{Attacks} & \multicolumn{2}{c}{Original} & \multicolumn{2}{c}{SelfSup+Con} & \multicolumn{2}{c}{SelfSup+Sup} & \multicolumn{2}{c}{Con+Sup} & \multicolumn{2}{c}{All}  \\ \cline{3-12} 
 &  & ABSE$\downarrow$ & $\delta\uparrow$ & ABSE$\downarrow$ & $\delta\uparrow$ & ABSE$\downarrow$ & $\delta\uparrow$ & ABSE$\downarrow$ & $\delta\uparrow$ & ABSE$\downarrow$ & $\delta\uparrow$ \\ \midrule\midrule
\multirow{8}{*}{\rotatebox[origin=c]{90}{Monodepth2}} 
 & L0 1/20 &4.711 &0.652  &\textbf{0.182}  &\textbf{0.998} & 0.193 & 0.989  & 0.529 &0.917 & \underline{0.187} & \underline{0.998} \\
 & L0 1/10 &6.088 &0.516  & \underline{0.244} &\underline{0.991} & 0.256& 0.988 &0.849  &0.761 & \textbf{0.24} &\textbf{0.991} \\
 & L0 1/5 &8.83 &0.393  & \underline{0.357} & \underline{0.967} & 0.394& 0.969 &1.293  &0.714 &  \textbf{0.288}&\textbf{0.986} \\
 & L0 1/3 &9.996 &0.344  &\underline{0.491}  &\underline{0.958} & 0.568& 0.954 &1.803  &0.667 & \textbf{0.419} &\textbf{0.97}\\
 & PGD &4.747 &0.56  & \textbf{0.681} & \textbf{0.988}&0.816& 0.982  &1.666  &0.72 & \underline{0.709} & \underline{0.986}\\
 & PGD & 11.684&0.343  & \textbf{1.332} &\underline{0.869} & \underline{1.586}&0.839  & 2.832 &0.646 &1.637  &\textbf{0.884}\\
 & PGD &17.109 &0.233  &\underline{3.382}  &0.591 & 3.546&\textbf{0.65}  &4.612  &0.472 &\textbf{3.056}  &\underline{0.621}\\
 & Patch & 2.714&0.778  & 0.758 & 0.945& \underline{0.331}& \underline{0.977} & 1.106 &0.798 &\textbf{0.277}  &\textbf{0.995}\\ \hline
\end{tabular}
\begin{tablenotes}
\item *Bold indicates the best performance in each row and underlining means the second best.
\end{tablenotes}
    \end{threeparttable}
}
\vspace{-5pt}
\end{table*}

\vspace{-7pt}
\subsubsection{Transfer to other target objects}\label{append:target_objs}

The target object under attack may not be used in training. In this experiment, we evaluate how the adversarially trained models perform on unseen target objects. We evaluate with four vehicles and one traffic barrier using the original and the three hardened Monodepth2 models and we perform $L_0$-norm-bounded attack on each object with $\epsilon=1/10$. 
Figure~\ref{fig:target_objects} shows the examples of target objects that we used in our transferability evaluation. Figure~\ref{fig:target_objects}(a) is the object we used in adversarial training and others are target objects under attack in the robustness evaluation. 
Table~\ref{tab:target_obj} shows the result. The first column is the target objects under attack and the following columns are the performance of different models. The first object (BMW SUV Black) is an object used in training and our hardened models are most robust on it. The others are unseen objects during training and the hardened models can still mitigate the attack effect a lot compared with the original model, which validates the transferability of our hardened models. \texttt{L0+SelfSup} still has the best performance in general. 

\begin{table}[t]
    \centering
    \caption{\small Comparison between supervised baseline trained with pseudo depth label (\texttt{L0+Sup(Pseudo)}) and ground-truth depth label (\texttt{L0+Sup(GT)}).}
    \label{tab:pseudo_vs_GT}
    \vspace{-5pt}
    \scalebox{0.9}{
    \begin{tabular}{ccc|cc|cccc}
    \toprule
       \multirow{2}{*}{Attacks}  &\multicolumn{2}{c}{Original} & \multicolumn{2}{c}{\makecell{L0+SelfSup\\(Ours)}}& \multicolumn{2}{c}{\makecell{L0+Sup\\(Pseudo)}} & \multicolumn{2}{c}{\makecell{L0+Sup\\(GT)}} \\ \cline{2-9}
 &ABSE         & $\delta\uparrow$ &ABSE         & $\delta\uparrow$ & ABSE         & $\delta\uparrow$         & ABSE           & 
$\delta\uparrow$           \\ \midrule\midrule
L0 1/10 & 6.08 & 0.51 & \textbf{0.25} &\textbf{0.98} &        0.94     &      {0.93}           &    {0.75}            &   0.92                 \\ 
L0 1/5 & 8.83 & 0.39 & \textbf{0.34} & \textbf{0.9} &         1.59      &       {0.85}           &     {0.98}           &      0.83              \\ 
PGD 0.1 &  11.68 & 0.34 & \textbf{1.53} & \textbf{0.85} &          2.53    &       { 0.71}          &     {2.44}           &       0.68             \\ 
PGD 0.2 & 17.1 & 0.23 & \textbf{3.46} & \textbf{0.69} &           6.14   &        {0.50}          &     {5.11}           &       0.32            \\ \bottomrule
\end{tabular}
}
\vspace{-10pt}
\end{table}

\begin{figure}[t]
    \centering
    \includegraphics[width=0.95\columnwidth]{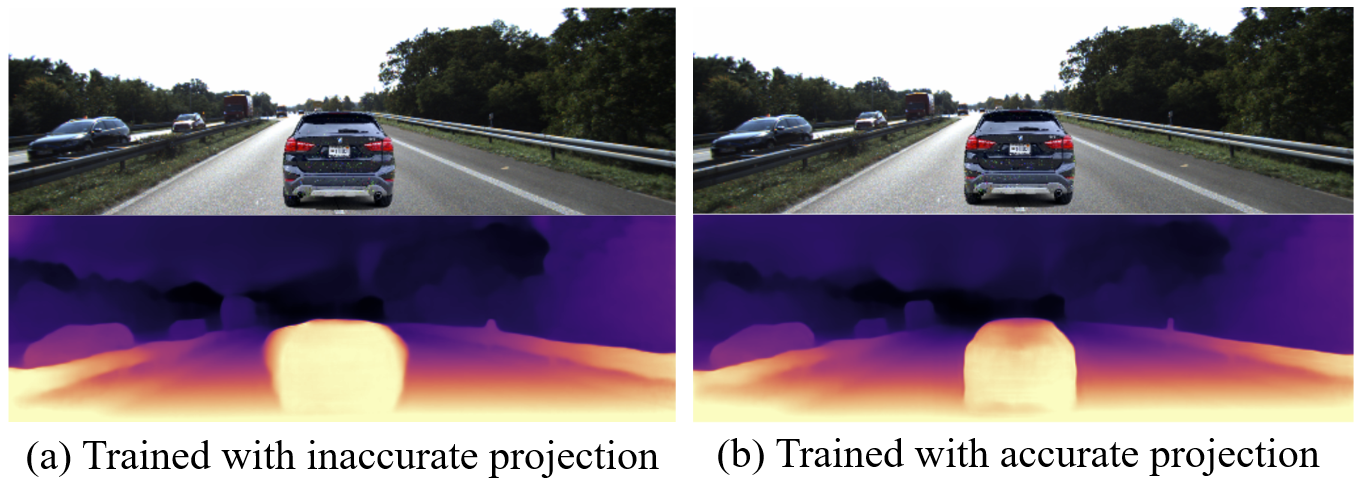}
    \vspace{-5pt}
    \caption{\small \textbf{Influence of inaccurate projection}. The outline of the depth estimation of the target object is blurred.}
    \label{fig:inacc_proj}
    \vspace{-10pt}
\end{figure}

\vspace{-5pt}
\subsubsection{Robustness against more attacks}\label{app:more_atks}
In addition to the attacks evaluated in our main paper (PGD attacks, $l_0$ attacks, adversarial patch attack and physical-world optimal patch attack), we have also evaluated the robustness of our models against more attacks: the $L_2$-bounded PGD attack~\cite{madry2018towards}, APGD attacks in AutoAttack~\cite{croce2020reliable}, Square attack~\cite{andriushchenko2020square}(a query-based black-box attack), Gaussian Blur~\cite{rauber2020foolbox} and AdvLight~\cite{duan2021adversarial}. The subject network is Mondepth2. Results can be found in Table~\ref{tab:perf_more_atks}. As shown, Our model is still more robust than the original model in all cases, and our method outperforms all others. Attacks with an $L_\infty$-bound of 0.1 can only cause less than 1 m depth estimation error on our hardened models while more than 10 m on the original model. The black-box attack does not have a good performance in the MDE task and only causes a mean depth estimation error of 0.924 m with 5000 queries to the original model. Although the Gaussian Blur and AdvLight are more stealthy and natural attacks, they are not as as effective as other methods in the task of MDE (\ie, per-pixel-based regression task).

\vspace{-5pt}
\subsubsection{Training From Scratch}\label{app:scratch}

We also compare the robustness performance of training from scratch and fine-tuning an existing model with our self-supervised adversarial training method using $L_0$-norm-bounded perturbation. We evaluate on Monodepth2 and Figure~\ref{fig:finetune} shows the ABSE result. As shown, training from scratch can provide more robustness, especially for higher-level attacks. As for the benign performance ($\ie,$ depth estimation performance), the ABSE of the fine-tuned model is 2.16 while the ABSE of the model trained from scratch is 2.468, meaning that the model trained from scratch has slightly worse than benign performance.

\vspace{-7pt}
\subsubsection{Adversarial Training Methods Combination}\label{app:combi}

We introduced three adversarial training methods in 
\S\ref{sec:methods}
and evaluated them separately in our main experiments. In this experiment, we explore the method combinations. Specifically, we combine different loss terms together as a total loss in adversarial training and there are four combinations in total. The perturbation in training is $L_0$-norm-bounded with $\epsilon=1/10$ and other evaluation setups are the same as our main experiments. The depth estimation performance ($\ie,$ clean performance) of each model is shown in Table~\ref{tab:comb_benign}. Results show that models trained with the self-supervised method have equivalent performance as the original model and \texttt{Con+Sup} performs slightly worse than the original one. Table~\ref{tab:comb_perf} shows the robustness performance under attacks. As shown, combining self-supervised learning with contrastive learning could achieve better robustness than self-supervised learning itself, and combining all three methods further improves the robustness.

\vspace{-5pt}
\subsubsection{Supervised Baseline with Ground Truth Depth}\label{app:sup_GT}

Although in the scope of this paper, we discuss self-supervised scenarios and assume the ground-truth depth is not available in training, we still conduct experiments comparing the defensive performance of our supervised baseline trained with pseudo-ground truth and that trained with ground-truth depth. We use Monodepth2 as the subject model. The results in Table~\ref{tab:pseudo_vs_GT} show that using ground-truth depth (\texttt{L0+Sup(GT)}) has a similar performance to using pseudo ground truth (\texttt{L0+Sup(Pseudo)}), and they are still worse than our self-supervised approach. Hence, whether to use ground-truth depth is not the bottleneck, and our pseudo-supervised baseline is not a weak choice, which also has significant defensive performance against attacks.

\vspace{-5pt}
\subsubsection{\added{Influence of Inaccurate View Synthesis}}\label{app:inacc_proj}

\begin{figure}[t]
    \centering
    \includegraphics[width=0.9\columnwidth]{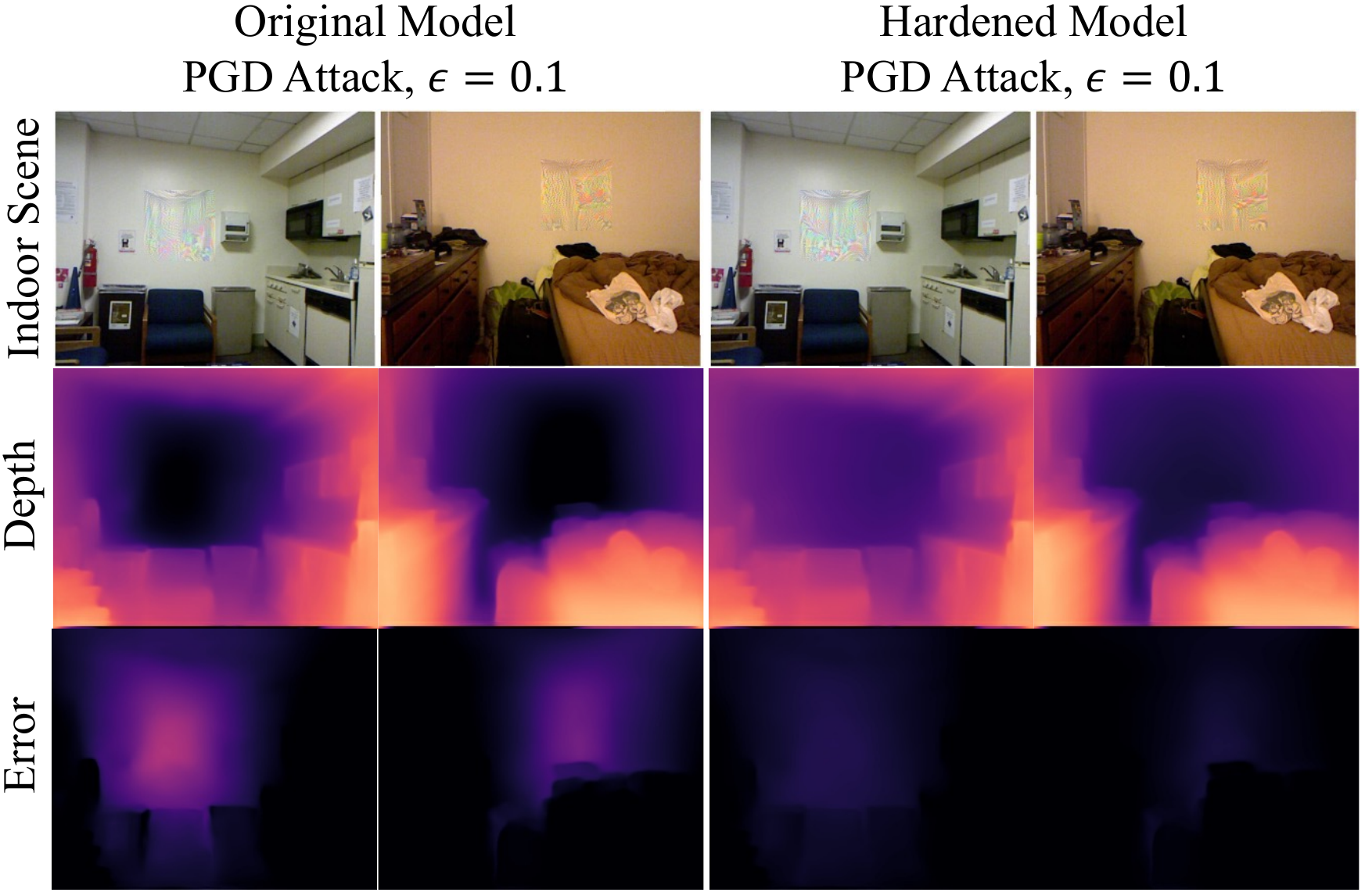}
    \vspace{-5pt}
    \caption{\small \addedblue{Qualitative examples of the defensive performance on \textbf{adversarial indoor scenes}.}}
    \label{fig:indoor_vis}
    \vspace{-10pt}
\end{figure}

\textbf{\added{Incorrect Projections.}} As stated in \S\ref{sec:view_syn}, we project a 2D image of the target object to two adjacent views considering the camera pose transformation $T_{t\rightarrow s}$ between $C_s$ and $C_t$ (Equation~\ref{eq:it_syn} and \ref{eq:is_syn}), then we enable the self-supervised training of MDE network by reconstructing $I_t$ from $I_s$ using the estimated depth $D_{I'_t}$ (Equation~\ref{eq:target_recons}). In this section, we explore what influence the inaccurate projections of the two camera views would have on training results. In this experiment, we only consider half of the camera pose transformation (\ie, $T_{t\rightarrow s}/2$) while projecting the object to $I_s$ instead of the true value $T_{t\rightarrow s}$, which leads to a wrong $I_s$ intentionally. For example, the distance between the stereo cameras in the KITTI dataset is 0.54 m, but we use 0.27 m to synthesize $I_s$. $I_t$ is still correctly synthesized. We use $L_0$-bounded perturbation with $\epsilon=1/10$ and self-supervised training to harden the Monodepth2 model. Figure~\ref{fig:inacc_proj} shows the result. As shown, the outline of the estimated depth of the target object is blurred when the MDE model is trained with inaccurate projection. It remains clear for the model trained with accurate projection.

\noindent\added{\textbf{Unrealistic Object Positions.} As stated in the training configurations in \S\ref{sec:ES}, in our main experiments, we did not exclude unrealistic images like objects on top of others from the synthesized training set because they are infrequent. In this section, we evaluate the influence of the unrealistic object positions on model hardening performance. In the new experiment, we employ a semantic segmentation model (Mask2Former~\cite{cheng2021mask2former}) on the sampled stereo image pairs first, then we exclude the synthesized images in which the target object is on top of others. We conduct adversarial training with Monodepth2. Results show that the model hardened with filtered images has similar defensive performance with models in our original setting. Specifically, the ABSE of the new model under $l_0$-norm-bounded attack with $\epsilon=1/10$ and PGD attack with $\epsilon=0.1$ are 0.248 and 1.535 respectively. In comparison, the ABSE of our previous model under those attacks are 0.251 and 1.533. The benign performance of the new model is slightly improved by 0.88\% with the ABSE being 2.141. Overall, it appears that the rare cases of unrealistic object positions in training data have minimal impact on performance.}

\vspace{-10pt}
\added{\subsubsection{Training without Object Synthesis}\label{app:without_obj}
As outlined in our introduction, we simulate the perspective shifts in real-world patch attack on objects by synthesizing objects onto stereo image pairs and adjusting the distance and viewing angles during view synthesis. In this section, we conduct ablation studies to assess how omitting our object synthesis technique affects model hardening performance. Differing from our initial approach, we generate adversarial perturbations directly onto $I^0_t$ of the stereo images taken from the dataset without synthesizing any target objects. Consistent with our usual protocol, we continue to use the proposed $L_0$-norm-bounded perturbation method (i.e., Equation~\ref{eq:perturb_gen}) with $\epsilon=1/10$, along with self-supervised adversarial training. All other training and evaluation configurations remain unchanged. Table~\ref{tab:without_syn} presents the defensive capabilities of models trained without object synthesis. The results indicate that these models have reduced defensive effectiveness on various attacks compared to those hardened using our standard method, confirming that our object synthesis technique plays a significant role in enhancing adversarial resilience.}


\begin{table}[t]
    \centering
    \caption{\added{\small Defensive performance of the model trained \textbf{without object synthesis.}} }
    \label{tab:without_syn}
    \vspace{-5pt}
    \scalebox{0.9}{
    \begin{tabular}{ccc|cc|cc}
    \toprule
       \multirow{2}{*}{Attacks}  &\multicolumn{2}{c}{Original} & \multicolumn{2}{c}{\makecell{L0+SelfSup\\(Ours)}}& \multicolumn{2}{c}{\makecell{L0+SelfSup\\(w/o Obj. Syn.)}}  \\ \cline{2-7}
 &ABSE $\downarrow$         & $\delta \uparrow$ &ABSE $\downarrow$         & $\delta \uparrow$ & ABSE $\downarrow$         & $\delta \uparrow$  \\ \midrule\midrule
L0 1/20 & 4.71 & 0.65 & \textbf{0.18} & \textbf{0.99} &        0.36     &      {0.98}               \\ 
L0 1/10 & 6.08 & 0.51 & \textbf{0.25 }&\textbf{ 0.98} &         0.82      &       {0.93}             \\ 
PGD 0.05 &  4.74 & 0.56 & \textbf{0.82} & \textbf{0.97} &          1.08    &       { 0.85}             \\ 
PGD 0.1 & 11.68 & 0.34 & \textbf{1.53 }& \textbf{0.85} &           2.17   &        {0.77}           \\ 
\bottomrule
\end{tabular}
}
\vspace{-10pt}
\end{table}

\vspace{-5pt}
\subsection{Extensions}\label{app:indoor-manydepth}

\subsubsection{Indoor Scenes}

Although we focus on outdoor scenarios like autonomous driving, a domain where self-supervised MDE is widely adopted (e.g., Tesla Autopilot), our technique hardens the MDE networks and does not have any assumptions about indoor or outdoor scenes. Actually, both Monodepth2 and DepthHints models have been proven to be directly applicable on indoor scenes (e.g., NYU-Depth-v2 Dataset) without any retraining~\cite{peng2021excavating,zhou2022devnet}, which implies the indoor-scene capability of our hardened models. We further conduct additional experiments with indoor scenes in the NYU-Depth-v2 Dataset. We launch adversarial attacks in a square area at the center of each scene to mimic a physical patch in the scene and evaluate the defensive performance of the original and hardened Monodepth2 models. Results are shown in Table~\ref{tab:nyu-dataset}. As shown, our hardened models reduce the depth estimation error caused by the attack significantly and the model trained with our self-supervised method has the best defensive performance. \addedblue{Qualitative examples are presented in Figure~\ref{fig:indoor_vis}. The first row illustrates adversarial indoor scenes, while the second and third rows depict depth estimation and errors induced by the adversarial examples, respectively. As demonstrated, our hardened model offers improved defense against adversarial attacks and reduces the error in depth estimation. } Therefore, the robustness of our hardened model still holds for indoor scenes.

\vspace{-5pt}
\subsubsection{Advanced Networks}

Monodepth2 and DepthHints are the two popular and representative self-supervised MDE networks, and they are widely used as baselines in the literature, so we use them as our subject networks. However, we do not have any assumptions about the MDE network structure, and our method should also work on state-of-the-art MDE models. Hence we conduct additional experiments with Manydepth~\cite{watson2021temporal}. We use our $L_0$-bounded perturbation with self-supervised adversarial training to harden the original Manydepth model, and Table~\ref{tab:manydepth-eval} shows the defensive performance of the original and hardened models. As shown, the original models are still vulnerable to both $L_0$-bounded and PGD attacks. The model hardened with our techniques is a lot more robust. The mean depth estimation error is reduced by over 80\%, which validates that our techniques are generic and also work on state-of-the-art MDE models.

\begin{table}[t]
    \centering
    \caption{\small Defensive performance on the \textbf{indoor scenes} with the NYU-Depth-v2 Dataset.}
    \vspace{-1mm}
    \label{tab:nyu-dataset}
    \scalebox{0.85}{
    \begin{tabular}{lcccccccc}
\toprule
\multicolumn{1}{c}{\multirow{2}{*}{Attacks}} & \multicolumn{2}{c}{Original} & \multicolumn{2}{c}{\makecell{L0+SelfSup \\(Ours)}} & \multicolumn{2}{c}{L0+Sup} & \multicolumn{2}{c}{L0+Contras} \\ \cline{2-9} 
\multicolumn{1}{c}{} & ABSE & $\delta \uparrow$ & ABSE & $\delta \uparrow$ & ABSE & $\delta \uparrow$ & ABSE & $\delta \uparrow$ \\ \midrule\midrule
L0 1/20 & 1.243 & 0.772 & \textbf{0.099} & \textbf{0.993} & 0.363 & 0.976 & 0.695 & 0.812 \\
L0 1/10 & 3.866 & 0.54 & \textbf{0.164} & \textbf{0.985} & 0.49 & 0.924 & 1.12 & 0.733 \\
PGD 0.05 & 1.784 & 0.727 & \textbf{0.12} & \textbf{0.986} & 0.456 & 0.849 & 1.641 & 0.717 \\
PGD 0.1 & 4.598 & 0.426 & \textbf{0.288} & \textbf{0.912} & 0.962 & 0.775 & 4.779 & 0.42 \\ \bottomrule
\end{tabular}
}
\end{table}

\begin{table}[t]
    \centering
    \caption{\small Defensive performance of the original and hardened \textbf{Manydepth models}.}
    \label{tab:manydepth-eval}
    \vspace{-5pt}
    \scalebox{0.9}{
    \begin{tabular}{lcccc}
\toprule
\multicolumn{1}{c}{\multirow{2}{*}{Attacks}} & \multicolumn{2}{c}{Original} & \multicolumn{2}{c}{L0+SelfSup} \\ \cline{2-5} 
\multicolumn{1}{c}{} & ABSE$\downarrow$ & $\delta\uparrow$ & ABSE$\downarrow$ & $\delta\uparrow$ \\ \midrule\midrule
L0 1/20 & 1.554 & 0.78 & \textbf{0.221} & \textbf{0.996} \\
L0 1/10 & 2.684 & 0.662 & \textbf{0.301} & \textbf{0.994} \\
PGD 0.05 & 7.112 & 0.417 & \textbf{1.270} & \textbf{0.855} \\
PGD 0.1 & 9.452 & 0.339 & \textbf{1.614} & \textbf{0.83} \\ \bottomrule
\end{tabular}
}
\vspace{-10pt}
\end{table}

\vspace{-5pt}
\section{Broader Impact and Limitations}\label{app:limit}

Our adversarial training method hardens the widely used monocular depth estimation networks and makes applications like autonomous driving more secure with regard to adversarial attacks. Compared to original training, the hardening of such models with our method does not require additional data or resources and the computational cost is affordable. The adversarial attacks we studied are from existing works and we do not pose any additional threats. Some limitations we could think of about our method are as follows. In our synthesis of different views, we assume the physical-world object is a 2D image board instead of a 3D model to avoid using an expensive scene rendering engine or high-fidelity simulator and to improve efficiency, which could induce small errors in synthesis though. However, since most physical-world attacks are based on adversarial patches attached to a flat surface, this 2D board assumption is realistic and practical. Precise synthesis should consider lighting factors such as glare, reflections, shadows, etc., but how to do that at a low-cost is an open problem and we leave it as our future work. In addition, although our modeling of the relative positions and viewing angles of the camera and physical object does not cover all real-world conditions, it considers the most common situations. There might be some corner cases in which the adversarial attack could escape from our defense, but we still mitigate the threats in the most common situations and improve the overall security a lot. \addedbrown{Additionally, our methodology is primarily designed to counteract adversarial attacks that exploit meticulously crafted adversarial noise to undermine the integrity of DNN-based MDE models. These adversarial perturbations are finely tuned to resonate with the target model, thereby facilitating the attainment of the attack goal. Our approach is designed to enable the model to disregard these adversarial noises during its prediction process. Given that the variations or distortions introduced in the image, such as those caused by a droplet of water on the camera lens, do not constitute the optimised adversarial noise aimed at a specific attack goal, they fall outside the purview of our defense mechanism and, as such, may not be effectively mitigated by our proposed solution. We leave it as a future direction to explore.}

\vspace{-5pt}
\section{Conclusion}
We address the issue of enhancing the resilience of self-supervised Monocular Depth Estimation (MDE) models against real-world attacks  without the necessity for depth ground truth data. Our strategy entails a self-supervised adversarial training process, leveraging view synthesis in conjunction with camera poses. Additionally, we incorporate $L_0$-norm-limited perturbation creation during the training phase to boost the model's robustness to attacks in the physical world. When benchmarked against conventional supervised learning and contrastive learning techniques, our method demonstrates superior resilience to a range of adversarial attacks, both in digital and physical environments, while maintaining virtually unchanged performance.

\noindent \textbf{Acknowledgement.} This research was supported, in part, by IARPA TrojAI W911NF-19-S-0012, NSF 1901242,  1910300 and 2242243, ONR N000141712045, N000141410468, and N000141712947. The views and conclusions contained herein are those of the authors and should not be interpreted as necessarily representing the official policies or endorsements, either expressed or implied, of the U.S. Naval Research Laboratory (NRL) or the U.S. Government.
			
{\small
    \bibliographystyle{IEEEtran}
    \bibliography{egbib}
}
			
\begin{IEEEbiography}[{\includegraphics[width=1in,height=1.25in,clip,keepaspectratio]{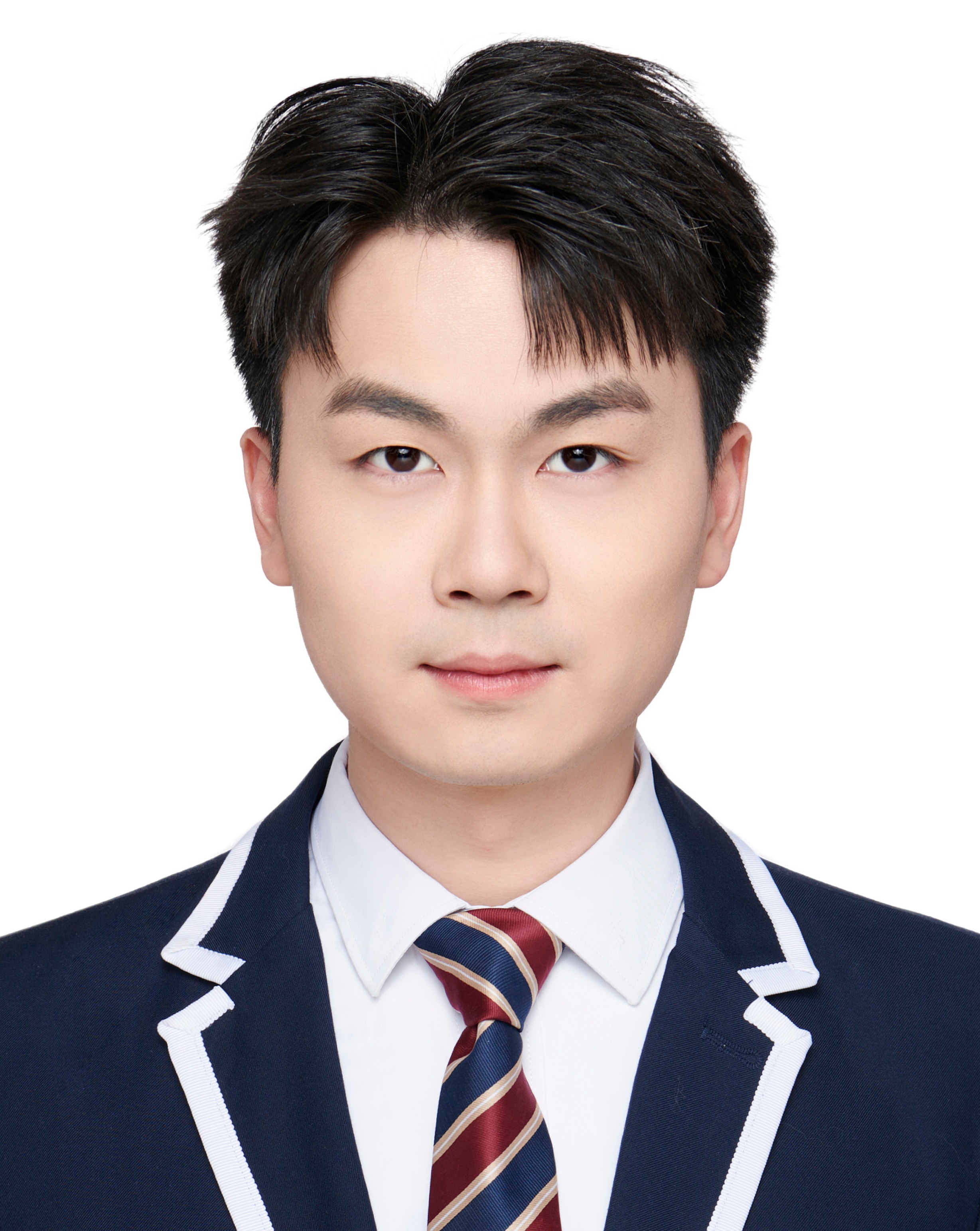}}]
{Zhiyuan Cheng} is currently a Ph.D. student at Purdue University, and he received his bachelor's degree in Computer Science and Technology from Wuhan University in 2019. His research interest lies at the intersection of AI and security, and he specializes in trustworthy machine learning. Specifically, he focuses on enhancing the robustness and security of AI-driven applications. He has publications in both top-tier security and machine learning conferences, such as NDSS, ICLR, ICML and ECCV.
\end{IEEEbiography}
\vspace{-6mm}
\begin{IEEEbiography}[{\includegraphics[width=1in,height=1.25in,clip,keepaspectratio]{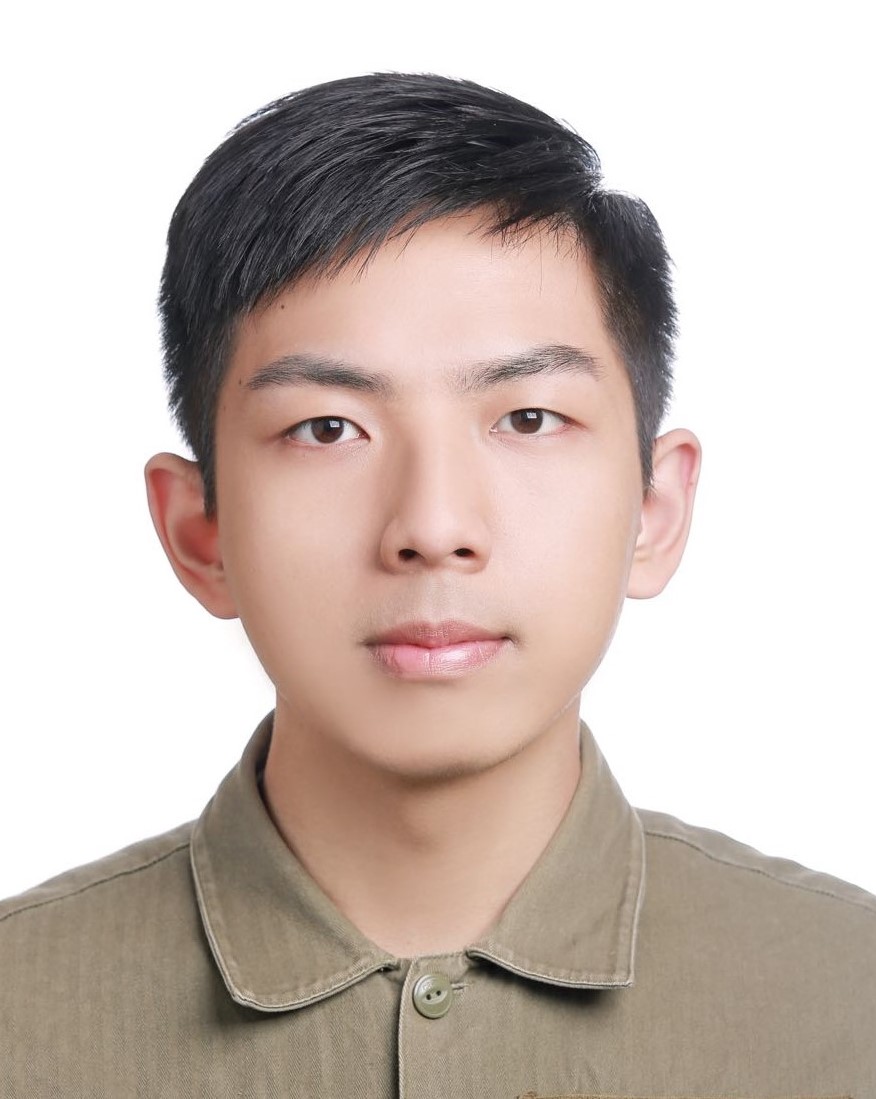}}]
				{Cheng Han} received his bachelor's degree from Tianjin University (TJU) in 2019, and his master's degree from the Pennsylvania State University (PSU) in 2021. He is currently a Ph.D. candidate at the Rochester Institute of Technology (RIT), and an incoming assistant professor in the School of Science and Engineering at the University of Missouri -- Kansas City (UMKC). His research interests include computer vision, deep learning, and adaptable and sustainable intelligence.
			\end{IEEEbiography}
\vspace{-6mm}
\begin{IEEEbiography}[{\includegraphics[width=1in,height=1.25in,clip,keepaspectratio]{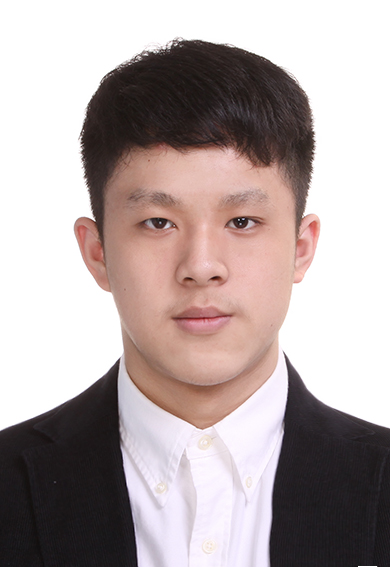}}]
{James Liang} received his bachelor's degree in Mathematics and Statistics from the University of Illinois at Urbana-Champaign, in 2021. He is currently a Ph.D. candidate in the Department of Computer Engineering at the Rochester Institute of Technology and a student trainee at the U.S. Naval Research Laboratory. His research interests include computer vision, deep learning, and machine learning.
\end{IEEEbiography}
\vspace{-6mm}
\begin{IEEEbiography}[{\includegraphics[width=1in,height=1.25in,clip,keepaspectratio]{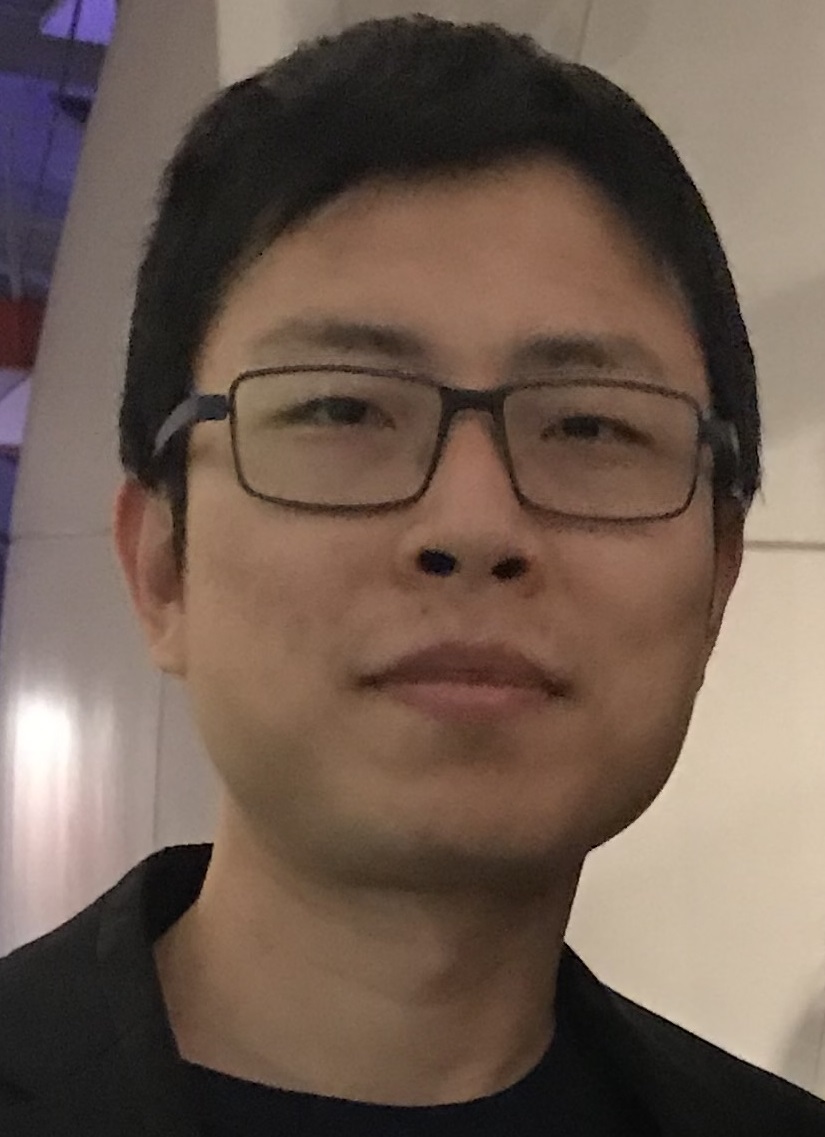}}]
{Qifan Wang} received his Ph.D. degree in computer science from Purdue University in 2015. He is currently a Senior Staff Research Scientist at Meta AI, leading a team building innovative Deep Learning and Natural Language Processing models for Recommendation Systems. He has co-authored over 60 publications in top-tier conferences and journals, including NeurIPS, SIGKDD, WWW, SIGIR, AAAI, IJCAI, ACL, EMNLP, WSDM, CIKM, ECCV, TPAMI, TKDE, and TOIS.
\end{IEEEbiography}
\vspace{-6mm}
\begin{IEEEbiography}[{\includegraphics[width=1in,height=1.25in,clip,keepaspectratio]{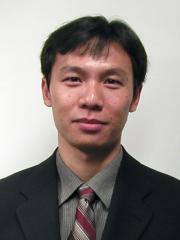}}]
{Xiangyu Zhang} is a Samuel Conte Professor at Purdue University, West Lafayette. He received his Ph.D. degree in computer science from the University of Arizona, Tucson, in 2006. His current research interest lies in AI security, software security, program analysis, and software engineering. He has published over 80 top-tier publications, including NDSS, USENIX Security, S\&P, CCS, ICML, \etc. He has received NSF Career Award and a few Best Paper Awards from top conferences as well.
\end{IEEEbiography}
\vspace{-6mm}
\begin{IEEEbiography}[{\includegraphics[width=1in,height=1.25in,clip,keepaspectratio]{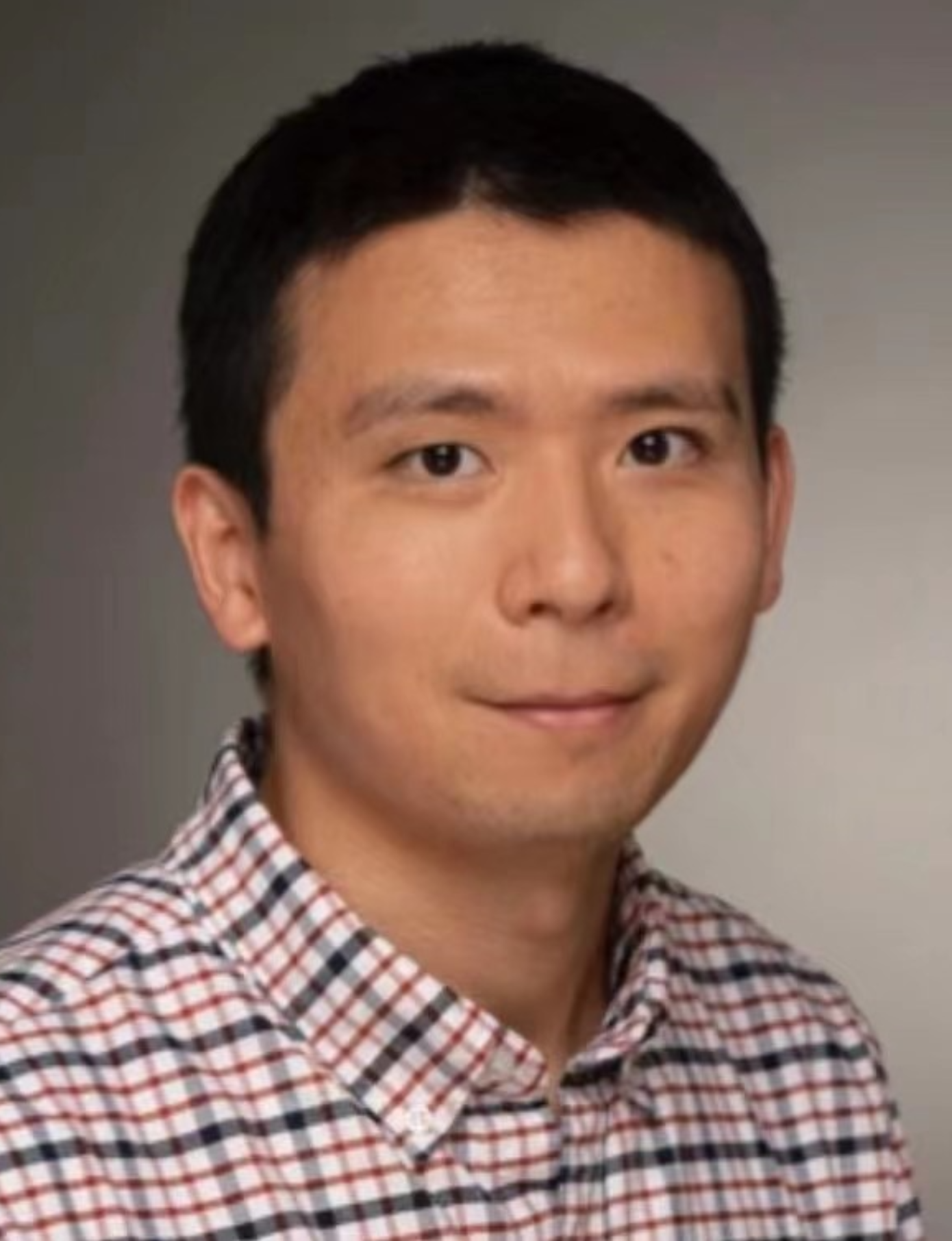}}]
{Dongfang Liu} is an assistant professor in the Department of Computer
Engineering at the Rochester Institute of Technology (RIT). He earned his
Ph.D. degree from Purdue University. Dr. Liu has dedicated his research efforts
to developing AI-based solutions for interdisciplinary research to address
challenges with societal relevance. His publication portfolio includes flagship
conferences in the AI and robotics domains, such as CVPR, ICCV, ECCV,
AAAI, IJCAI, WACV, WWW, and IROS.
\end{IEEEbiography}
		\end{document}